\documentclass{vgtc}      

\graphicspath{{figures/}{pictures/}{images/}{./}} 
\usepackage[table]{xcolor}
\usepackage{times}                     
\usepackage{tabu}                      
\usepackage{booktabs}                  
\usepackage{lipsum}                    
\usepackage{mwe}                       
\usepackage{orcidlink}
\usepackage{amsmath, amssymb}
\usepackage[capitalize]{cleveref}
\crefname{section}{Sec.}{Secs.}
\Crefname{section}{Section}{Sections}
\Crefname{table}{Table}{Tables}
\crefname{table}{Tab.}{Tabs.}
\crefname{figure}{Fig.}{Figs.}
\crefname{algorithm}{Alg.}{Algs.}
\usepackage{algorithm}
\usepackage{algorithmic}
\usepackage{amsmath}
\usepackage{multirow}
\usepackage{adjustbox}
\usepackage{mathptmx}                  
\usepackage[flushleft]{threeparttable}

\onlineid{1046}

\vgtccategory{Research}

\title{
High-Fidelity Mask-free Neural Surface Reconstruction 
for Virtual Reality}

\author{Haotian Bai\thanks{email: haotianwhite@outlook.com}\\ %
\scriptsize HKUST(GZ) %
\and Yize Chen \thanks{email: yize.chen@ualberta.ca}\\ %
\scriptsize University of Alberta %
\and Lin Wang\thanks{Corresponding author, email:  linwang@ust.hk}\\ %
     \parbox{1.4in}{\scriptsize \centering HKUST(GZ) \\ HKUST}}
 \vspace{-15pt}
\teaser{
  \centering
  \vspace{-12pt}
  {Project homepage: \url{https://vlislab22.github.io/Hi-NeuS/}}
  \includegraphics[width=\linewidth]{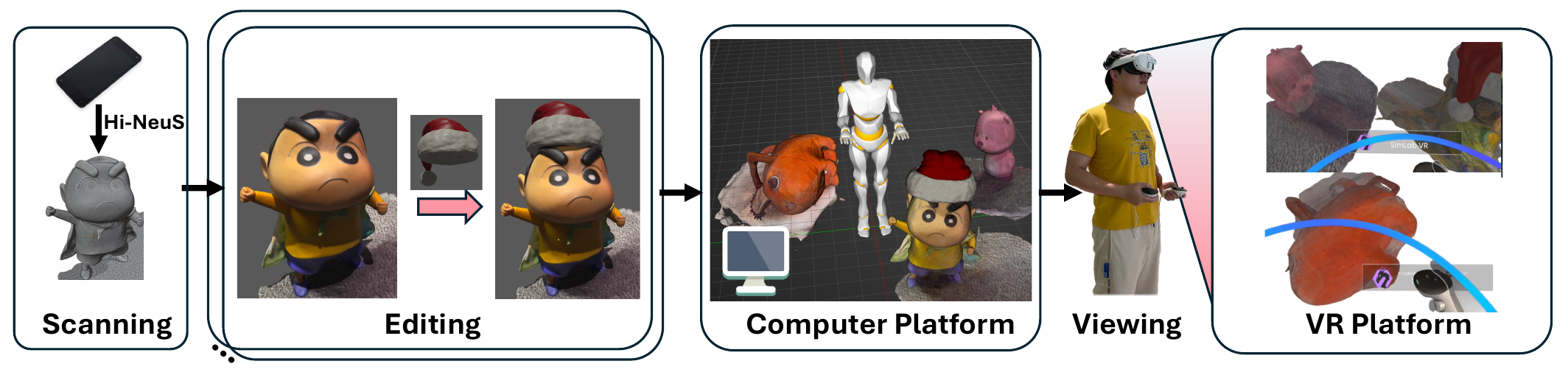}
  \vspace{-10pt}
  \caption{\textbf{
Overview of our Hi-NeuS Framework:} 
It converts multi-view images captured from phone cameras into digitized mesh objects. The meshes can be edited and integrated into both computer and VR for viewing and editing. This framework can be potentially employed for direct geometry capturing and subsequent viewing and creation in VR/AR. 
  \label{fig:teaser}
}
}

\abstract{
Object-centric surface reconstruction from multi-view images plays a crucial role in creating editable digital assets for AR/VR.
Due to the lack of geometric constraints, existing methods, \textit{e.g.}, NeuS~\cite{Wang_Liu_Liu_Theobalt_Komura_Wang_2021} necessitate annotating the object masks to reconstruct compact surfaces in mesh processing.
Mask annotation, however, is labor-intensive due to its cumbersome nature,
and its absence may lead to noisy surfaces. 
This paper presents \textbf{Hi-NeuS}, a novel rendering-based framework for neural implicit surface reconstruction, aiming to recover \textit{compact and precise surfaces without multi-view object masks}.
Our key insight is that the overlapping regions in the object-centric views naturally
highlight the object of interest as the camera orbits around
objects. 
The object of interest can be essentially specified by estimating the distribution 
of the rendering weights accumulated from multiple views, which implicitly identifies the surface that a user intends to capture.
This inspires us to design a geometric refinement approach, which takes multi-view rendering weights to guide the signed distance functions (SDF) of neural surfaces in a self-supervised manner.
Specifically, it retains these weights to resample a pseudo surface based on their distribution. This facilitates the alignment of the SDF to the object of interest. We then regularize the SDF's bias for geometric consistency. 
Moreover, we propose to use unmasked Chamfer Distance(CD) to measure the extracted mesh without post-processing for more precise evaluation.
Our approach's effectiveness has been validated through NeuS and its variant Neuralangelo, demonstrating its adaptability across different NeuS backbones.
Extensive 
benchmark on the DTU dataset shows that our method reduces surface noise by about 20\%, and improves the unmasked CD by around 30\%, meanwhile, achieving better surface details. The superiority of Hi-NeuS is further validated on the BlendedMVS dataset, as well as the real-world applications using \textit{handheld} camera captures for content creation. 
} 


\begin{document}


\firstsection{Introduction}

\maketitle
%
%
%
Imagine the last time you captured photos of an object by walking around it. Now, you want to integrate that object into a 3D virtual environment, such as an AR/VR world, ideally in a mesh format for both viewing and content creation, as depicted at \cref{fig:teaser}. 
Traditionally, this process relied on classical stereo-based methods~\cite{Seitz_Dyer_2002, Tola_Strecha_Fua_2012, Andrew_2001, Laurentini_1994, Agrawal_Davis_2005, Furukawa_Ponce_2007, Schönberger_Zheng_Frahm_Pollefeys_2016}. However, recent advancements in 3D reconstruction using neural volume rendering~\cite{Mildenhall_Srinivasan_Tancik_Barron_Ramamoorthi_Ng_2020, Mescheder_Oechsle_Niemeyer_Nowozin_Geiger_2019} have transformed it, enabling the recovery of high-fidelity details and more complex structures. 
%
Compared to explicit representations like Gaussian Splatting~\cite{dai2024highquality, huang20242d, yu2024gaussian, Kerbl20233DGS}, which consists of dense collections of 3D Gaussians, the implicit NeRF is commonly used for surface reconstruction due to its maturity in terms of conversion and compatibility, as demonstrated by industry practices~\cite{zgroup2023revolutionizing}.
Neural implicit surface reconstruction typically employs multi-layer perceptrons (MLPs) to implicitly represent scenes as occupancy fields~\cite{Oechsle_Peng_Geiger_2021}, SDF\cite{Yariv_Gu_Kasten_Lipman_2021, Wang_Liu_Liu_Theobalt_Komura_Wang_2021}, or hybrid grids\cite{Neuralangelo}. Due to the inherent continuity of implicit representations, these methods can synthesize plausible novel view images. 
However, they lack sufficient surface constraints and struggle to extract high quality surfaces~\cite{Oechsle_Peng_Geiger_2021}.

\begin{figure*}[t!]
    \centering
    \includegraphics[width=\linewidth]{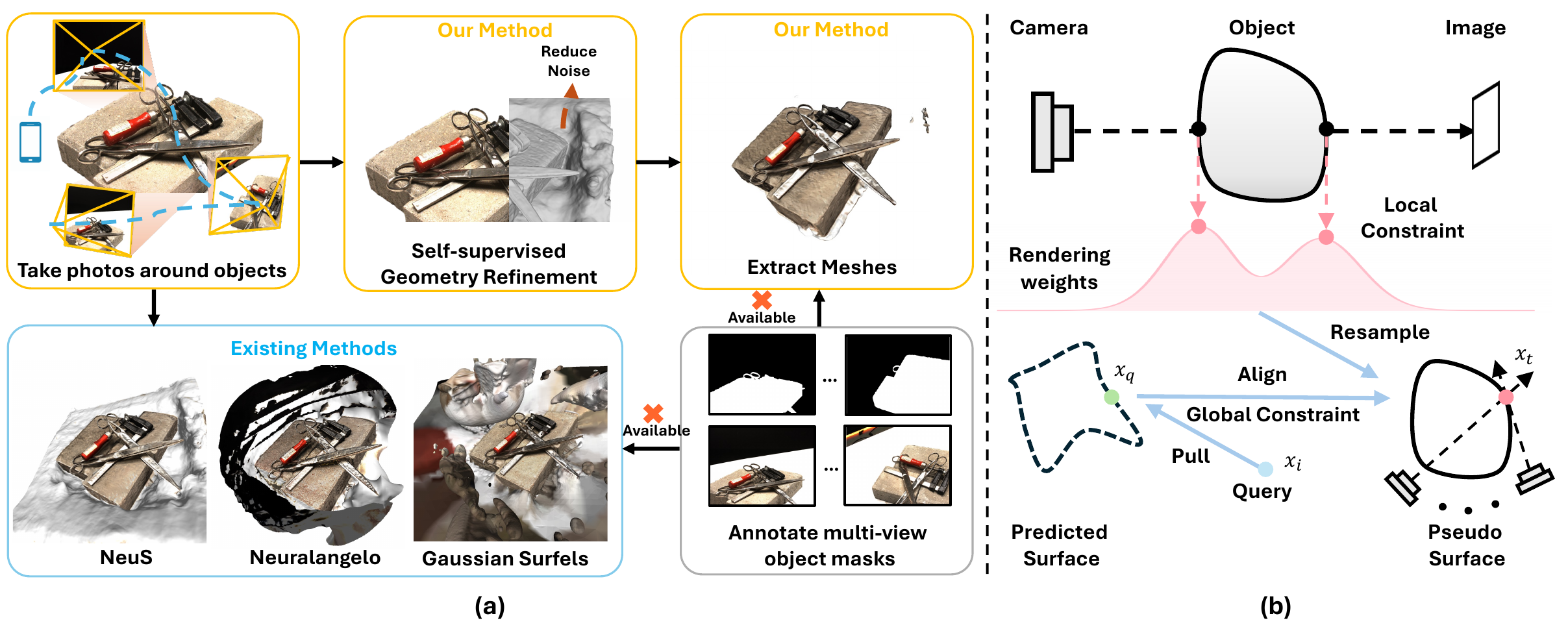}
    \vspace{-16pt}
    \caption{
   \textbf{(a) Comparison on surface reconstruction without masks. } We aim to reduce the noise in surface reconstruction without relying on multi-view object masks. Compared to existing methods, including NeuS~\cite{Wang_Liu_Liu_Theobalt_Komura_Wang_2021}, Neuralangelo~\cite{Neuralangelo}, and Gaussian Surfels~\cite{dai2024highquality}, Hi-NeuS produces compact and more precise mesh results, enhancing its utility for downstream applications in virtual reality.
 \textbf{(b) Self-supervised geometry refinement. } The rendering weights from multiple views are accumulated, corresponding to resampled target surface points $x_t$(\textcolor{red}{red}). Based on this supervision, we advance query points $x_i$ (\textcolor{blue}{blue}) to obtain the predicted surface points $x_q$ (\textcolor{teal}{green}). We then align them using Chamfer Distance (CD) with global geometric constraints related to SDF.
    }
    \label{fig:overview}
\end{figure*}

To tackle these issues, some works~\cite{Yariv_Gu_Kasten_Lipman_2021, Wang_Liu_Liu_Theobalt_Komura_Wang_2021, Oechsle_Peng_Geiger_2021} integrate implicit representations in volume rendering to reduce inherent geometry errors. Among them, NeuS~\cite{Wang_Liu_Liu_Theobalt_Komura_Wang_2021} is one of the pioneering works that adopt SDF-based volume rendering to model geometric surfaces. It integrates SDF into the density field in volume rendering to constrain the scene, yielding unbiased surface reconstruction with occlusion awareness. Notably, NeuS reduces the reliance on multi-view object masks as training supervision.
Despite NeuS’s superiority, its neural surface representation via SDF remains under-constrained.
Specifically, when the SDF is not fully trained, it struggles to accurately represent the underlying geometry, resulting in a biased SDF distribution~\cite{Chen_Zhang_Feldmann_Schreer_Eisert_2022}. This bias causes geometry errors, leading the predicted surface to deviate from the expected geometry as the noise to be reduced at the middle of (a) in \cref{fig:overview}.
%
%
%
Therefore, to extract compact mesh objects from learned SDF representations, NeuS and its follow-up works~\cite{Neuralangelo, Wang_Liu_Liu_Theobalt_Komura_Wang_2021, Wang_Han_Habermann_Daniilidis_Theobalt_Liu_2022} often require annotated masks for each camera pose during training or to remove background mesh noise in post-processing. However, the \textit{annotation process} is labor-intensive and prone to human error. 
Considering that all perspectives must be annotated, this approach becomes increasingly cumbersome and costly.
In scenes with complex or overlapping objects, these masks frequently struggle to delineate boundaries accurately, even when using methods like the Segment Anything Model(SAM)~\cite{kirillov2023segment}. 
On the other hand, direct filtering with the annotation masks is not plausible, as it leads to geometry artifacts, such as jagged edges or holes as depicted \cref{fig:eval_pipe}. 
We refer readers to the further discussion in our \textit{suppl.} materials.

%
%

Despite recent efforts to enhance geometric accuracy, many approaches continue to prioritize improving neural representation power~\cite{Darmon_Bascle_Devaux_Monasse_Aubry_2022} or assist surface reconstruction through auxiliary point cloud supervision~\cite{Zhang_Yao_Li_Fang_McKinnon_Tsin_Quan_2022} and pretrained models~\cite{Chen_Lu_Feldmann_Schreer_Eisert_2023}. The challenge of surface reconstruction using solely multi-view images, without multi-view object masks, remains under-explored. As shown in (a) of \cref{fig:overview}, the absence of these masks can lead to significant mesh artifacts in existing methods. This makes it hard to achieve a direct mesh reconstruction such as the pipeline shown at \cref{fig:teaser}.


To recover more compact and precise surfaces without object mask, \textit{our key inspiration is that, as the camera orbits around objects, the overlapping regions in the captured views naturally highlight the object of interest}. This overlap implicitly identifies the subject the photographer intends to capture.
Similarly, during volume rendering, as illustrated in the upper part at (b) of \cref{fig:overview}, the rendering weights peak near the object’s surface when the camera rays intersect with it,
which delineates the surface boundaries.
However, this ray-wise local constraint alone is insufficient to obtain the surfaces due to the biased SDF distribution. To address this, we propose leveraging the accumulated weights from multiple views to more effectively capture the global surface regions.

This idea builds on the findings of NeuS~\cite{Wang_Liu_Liu_Theobalt_Komura_Wang_2021}, which demonstrates that, with an unbiased rendering weight function, surface points contribute more to their corresponding ray pixels than other ray samples.
Furthermore, occlusion awareness ensures that the first intersection along a ray holds a higher value than subsequent intersections. Compared with this local geometric constraint, accumulating multi-view rendering weights may potentially delineate the object’s surface with similar peak weights regardless of ray directions.
Given this surface supervision, we can mitigate the geometry bias introduced during training by aligning SDF globally towards the generated pseudo-surfaces from these rendering weights. 
Specifically, as the geometric refinement process depicted at the lower part at (b) of \cref{fig:overview}, for a given arbitrary ray query,  upon the inference of SDF via MLPs, we employ the differentiable neural pulling operation~\cite{Baorui_Han_Liu_Zwicker_2020} to pull the ray sample towards the object's surface, guided by the predicted signed distances and their gradients. This approach allows gradients from these predicted surface points to be back-propagated into the neural SDF.
Hi-NeuS aligns these points with the target surface supervision,  
which can be resampled based on the accumulated multi-view weights distribution to locate the underlying surface. Thus, this supervision serves as the reference, imposing geometric constraints on the SDF to approximate a globally consistent surface through a self-supervised methodology. 
%
%
%
%
%
%
%
%

Our key contributions are as follows:
\textbf{(1)} We propose Hi-NeuS to create more precise and compact surfaces without requiring multi-view object mask annotations. This framework can potentially be used for direct geometry recovery; subsequent viewing and content creation in VR/AR.
\textbf{(2)} We propose using a geometric constraint to regularize the surface globally. This method leverages rendering weights accumulated from multiple views to align with the surface that a user intends to capture.
\textbf{(3)}
Comprehensive experiments using the NeuS backbone on the DTU dataset~\cite{Jensen_Dahl_Vogiatzis_Tola_Aanaes_2014} and the BlendedMVS dataset~\cite{Yao_Luo_Li_Zhang_Ren_Zhou_Fang_Quan_2020} demonstrate that our method significantly reduces noise while enhancing geometric detail. Additionally, our method’s versatility is validated across NeuS and its variant Neuralangelo~\cite{Neuralangelo} and exhibits superior performance in challenging \textit{real-world} scenarios involving handheld phone camera captures. 
%

\section{Related Works}
\noindent\textbf{3D content creation and interaction.}
Advances in neural rendering and real-time graphics have significantly enhanced 3D content viewing and interaction in AR/VR, enabling more immersive and interactive experiences across devices.
Recent technologies, such as Re-ReND~\cite{rojas2023rerendrealtimerenderingnerfs} and RT-NeRF~\cite{li2022rtnerfrealtimeondeviceneural}, have demonstrated real-time rendering of NeRFs in VR/AR headsets using standard graphics pipelines, offering high-quality visual experiences. Additionally, FoV-NeRF~\cite{deng2022fovnerffoveatedneuralradiance} improves rendering by focusing on the user’s gaze and optimizing computational resources. 
Furthermore, the VR-GS~\cite{jiang2024vrgsphysicaldynamicsawareinteractive} integrates physical dynamics for realistic, responsive interaction with 3D contents represented with Gaussian Splatting, ensuring a comprehensive virtual experience. 

Despite the rapid progress in integrating NeRF/GS into VR/AR, 
acquiring high-quality 3D assets remains a significant challenge in populating virtual worlds with 3D content.
Unlike generation methods~\cite{wang2022nerfarttextdrivenneuralradiance} that rely on text or images, which prioritize creativity, 3D reconstruction focuses on accuracy and realism. Consequently, the resulting meshes often exhibit intricate details and better align with the user's desired outcome.
To facilitate the process, several approaches have been explored, including few-shot novel view synthesis\cite{zhu2024cmcfewshotnovelview, long2022sparseneusfastgeneralizableneural}, more efficient NeRF representations\cite{yang2021recursivenerfefficientdynamicallygrowing}, and reducing training time~\cite{peng2023intrinsicngpintrinsiccoordinatebased}. In contrast, our approach focuses on recovering compact and precise surfaces without relying on object masks, thereby reducing the need for costly annotation.

\noindent\textbf{Surface reconstruction from multi-view images.}
Before the deep learning era, traditional multi-view stereo (MVS) methods dominated the field of surface reconstruction from multi-view images~\cite{Andrew_2001}. These techniques primarily reconstructed 3D shapes by matching features across adjacent frames~\cite{Kutulakos_Seitz_1999, Laurentini_1994, Agrawal_Davis_2005}, employing discretized frameworks like voxel grids~\cite{Laurentini_1994, Agrawal_Davis_2005}, and point clouds~\cite{Furukawa_Ponce_2007, Schönberger_Zheng_Frahm_Pollefeys_2016}. However, they often struggled to capture fine geometric details due to the limited resolution of their cost volumes.
The recent advent of Neural Radiance Fields (NeRFs)~\cite{Mildenhall_Srinivasan_Tancik_Barron_Ramamoorthi_Ng_2020} brings a paradigm shift with its continuous volumetric representation. NeRFs utilize an MLP to encode 3D scenes, 
correlating spatial locations with their corresponding colors and densities for photo-realistic volumetric rendering.
To further enhance implicit geometry, 
a variety of methods~\cite{Darmon_Bascle_Devaux_Monasse_Aubry_2022, Oechsle_Peng_Geiger_2021, Wang_Liu_Liu_Theobalt_Komura_Wang_2021} have been introduced. They aim to revise the rendering procedure to handle occlusions and sudden depth changes.  Additionally, other methods~\cite{Wang_Skorokhodov_Wonka_2022, Neuralangelo} focus on enhancing representational capabilities and training strategies to improve surface estimation accuracy.
Notably, Neuralangelo~\cite{Neuralangelo} proposes multi-resolution 3D hash grids with coarse-to-fine optimization integrated with SDF-based rendering, yielding state-of-the-art (SOTA) geometry accuracy and rendering capability.

On the other hand, Gaussian Splitting (GS) has gained significant attention for representing complex scenes using 3D Gaussians, with GS-based surface reconstruction demonstrating impressive performance compared to NeRF, particularly in terms of training and inference efficiency~\cite{dai2024highquality, huang20242d, yu2024gaussian, dai2024highquality}. However, NeRF offers more geometric detail while maintaining compactness and reducing overfitting, even when dealing with intricate geometries, textures, and material properties.
The adoption of NeRF for geometry has also been validated in recent industry practices~\cite{zgroup2023revolutionizing}, owing to its maturity in terms of conversion and compatibility.

Unlike previous approaches, our work aims to recover more compact and precise surfaces in neural implicit representations, eliminating the need for multi-view object masks. By doing so, we mitigate geometry artifacts, such as jagged edges and holes, which can potentially reduce the requirement for additional annotations.
Although recent advancements~\cite{Neuralangelo, Wang_Liu_Liu_Theobalt_Komura_Wang_2021, Oechsle_Peng_Geiger_2021, Yariv_Kasten_Moran_Galun_Atzmon_Basri_Lipman_2020} have enabled surface reconstruction without auxiliary object masks as training supervision, they may still rely on them to refine meshes, necessitating time-consuming annotation or manual editing to eliminate geometric noise. This limitation hinders the development of surface reconstruction methods with solely multi-view images.
%
%
%

\noindent\textbf{Geometrical constraints for neural representations.}
%
To represent the scene geometry, implicit functions such as occupancy grids~\cite{Oechsle_Peng_Geiger_2021, Niemeyer_Mescheder_Oechsle_Geiger_2020} and SDFs~\cite{Neuralangelo, Wang_Liu_Liu_Theobalt_Komura_Wang_2021} are preferred due to their continuous representation with low memory consumption. 
Recent works~\cite{Oechsle_Peng_Geiger_2021, Wang_Liu_Liu_Theobalt_Komura_Wang_2021, Darmon_Bascle_Devaux_Monasse_Aubry_2022} employ implicit functions to enforce geometric consistency across multi-view images, thereby imposing geometrical constraints on the learned object representation.
%
For instance, NeuS~\cite{Wang_Liu_Liu_Theobalt_Komura_Wang_2021} parameterizes the volume density and integrates it into the volume rendering process, achieving unbiased surface reconstruction with occlusion awareness. 
NeuralWarp~\cite{Darmon_Bascle_Devaux_Monasse_Aubry_2022} enhances geometry accuracy by warping views to learn from high-frequency image textures. Neuralangelo~\cite{Neuralangelo} enhances surface smoothness with continuous numerical gradients in its hash grids and predicted curvature.
However, these geometric constraints are limited to the image, ray, or patch level, imposing only regularization based on partial visual cues. This approach may lack the 3D spatial awareness needed for more consistent geometry regularization.

To mitigate this gap between 2D and 3D, follow-up works use auxiliary information to enhance global geometrical consistency. For instance, NeuralWarp~\cite{Darmon_Bascle_Devaux_Monasse_Aubry_2022} and RegSDF~\cite{Zhang_Yao_Li_Fang_McKinnon_Tsin_Quan_2022} leverage information from structure-from-motion to guide surface optimization. D-NeuS~\cite{Chen_Zhang_Feldmann_Schreer_Eisert_2022} utilizes depth maps to correct geometrical deviations in the SDF values at surface intersection points. Additionally, it employs a pre-trained model to enhance feature representation ability with RGB inputs. Similarly, Chen \textit{et al.}~\cite{Chen_Lu_Feldmann_Schreer_Eisert_2023} introduces a probability mask to refine pixel sampling on the foreground object, complemented by segmentation masks. 
The recent works using Gaussian Splatting for surface reconstruction, \textit{i.e.}, Gaussian Surfels~\cite{dai2024highquality} also rely on external object masks and surface norms to assist geometry learning during training.
These approaches' reliance on external models or data can be a considerable constraint when such resources are unavailable or their information is inaccurate, limiting their applicability to general cases.  
In contrast, our work introduces a self-supervised geometric refinement approach that leverages auto-generated rendering weights for surface refinement, establishing a global 3D geometrical constraint to regularize the geometry representation automatically. 
\section{Hi-NeuS Framework}
\subsection{Overview}

\begin{figure*}[t!]
    \centering
    \includegraphics[width=0.8\linewidth]{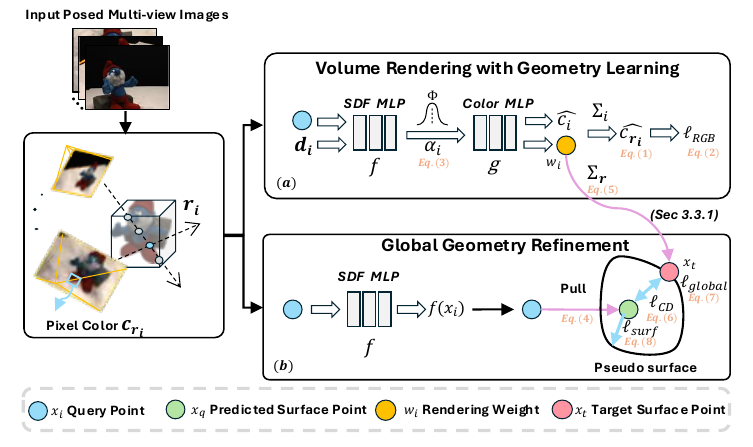}
    \caption{\textbf{Our proposed Hi-NeuS training framework}:
%
%
In volume rendering combined with geometry learning, we capture rendering weights from multiple views. Hi-NeuS then resamples based on the weight distribution to obtain supervisory surface points. Finally, global geometric refinement is applied using geometric constraints.
    }
    \label{fig:method}
\end{figure*}

As shown in \cref{fig:method}, given a set of posed multi-view images, surface reconstruction seeks to reconstruct the 3D object surfaces.
The Hi-NeuS training framework uses the same SDF-based volume rendering as our baseline NeuS~\cite{Wang_Liu_Liu_Theobalt_Komura_Wang_2021} and Neuralangelo~\cite{Neuralangelo}. This approach integrates a geometry representation SDF \( f(x) \) into a color MLP \( g(x) \) to generate images. Specifically, given an arbitrary point sample \( x_i \) and its corresponding ray direction $\mathbf{d_i}$, SDF-based volume rendering aims to convert SDF values into a volume density field $\alpha_i$ using the logistic function.
Then, the activated density with $\mathbf{d_i}$ is sent to $g(x)$ to infer the  color $\hat{\mathbf{c_i}}$. 
At last, it
predicts the corresponding pixel color \(\mathbf{\hat{C}(r_i)} \) by accumulating ray samples along rays supervised by the ground truth \( \mathbf{C(r_i)} \).
Throughout this process, Hi-NeuS records rendering weights \( w_i \) from multiple views.
Then \( w_i \) are transformed to form a probability distribution, from which target surface points \( x_t \) are resampled from the pseudo surface as supervisory signals.
The 
 predicted surface points \( x_q \) is obtained via the Neural Pulling operation~\cite{Baorui_Han_Liu_Zwicker_2020} to allow gradient updates to be back-propagated into the SDF. 
Finally, Hi-NeuS aligns \( x_q \) towards \( x_t \) in a self-supervised way, while regularizing geometric consistency. We refer readers to the algorithm's pseudo-code in our \textit{suppl.}.

\subsection{Preliminary: Volume Rendering with Geometry Learning}
As shown in \cref{fig:method}(a), we introduce the volume rendering process with input posed multi-view images to infer the rendering color and multi-view rendering weights. Those weights are then sent into the global geometric refinement to reduce geometric noise.

\noindent\textbf{Neural Volume Rendering.}
The neural volume rendering process involves learning the parameters of two implicit functions, $f(x)$ and $g(x)$. 
To infer the color $\hat{C}(\mathbf{r}_i)$ of a ray $\mathbf{r}_i$, we integrate $N$ samples $x_i$ along the ray. $\hat{C}(\mathbf{r}_i)$ for the corresponding pixel is computed by summing the weighted colors $\mathbf{\hat{c_i}}$ of the samples $x_i$ along the ray, where each sample $x_i$ is weighted by $w_i$:
\begin{align}
\label{eq:vol_rend}
\hat{C}(\mathbf{r_i}) = \sum_{i=1}^{N}w_i\hat{\mathbf{c_i}}, \; \text{where} \ w_i = T_i\alpha_i,
\end{align}
where \( \alpha_i = 1-\exp(-\sigma_i \delta_i) \) denotes the opacity of the \( i \)-th ray segment, and the light accumulated through any ray \( \mathbf{r_i} \) to sample \( i \) is represented by \( T_i = \prod_{j=1}^{i-1}(1-\alpha_j) \). Here, \( \sigma_i \) denotes the volume density, and \( \mathbf{c}_i \) is the color vector in the form of RGB or spherical harmonics (SH). The predicted color \( \hat{C}(\mathbf{r_i}) \) is optimized to approximate the ground truth (GT) color \( C(\mathbf{r_i}) \) by minimizing the mean squared error (MSE) loss:
\begin{align}
\label{eq:l_rgb}
\mathcal{L}_{RGB} = \sum_{i} \| C(\mathbf{r_i}) - \hat{C}(\mathbf{r_i}) \|_2^2
\end{align}

\noindent\textbf{Neural Rendering with SDF.}
The surface $\mathbb{S}$ of an SDF is defined by its zero-level set: $\mathbb{S} = \{x\in \mathbb{R}^3 | f(x)=0\}$. NeuS~\cite{Wang_Liu_Liu_Theobalt_Komura_Wang_2021} converts the SDF into volume density $\alpha$ using a logistic function $\phi_s(f(x))$, which is the derivative of the sigmoid function $\Phi_s$.
The discretized approximation of volume rendering is computed similarly to \cref{eq:vol_rend}, with the revised opacity given by:
\begin{align}
\label{eq:neus_opacity}
\alpha_i = \max \left( \frac{\Phi_s (f(x_i)) - \Phi_s (f(x_{i+1}))}{\Phi_s (f(x_i))}, 0 \right).
\end{align}

\noindent\textbf{Neural Pulling with neural SDF.}
To enhance the quality of SDF representations, Baorui et al.~\cite{Baorui_Han_Liu_Zwicker_2020} proposed a differentiable method that pulls a query 3D location \( x_i \) toward its closest surface intersection. As shown in \cref{fig:overview}, this predicted surface intersection is denoted as \( x_{q} \). It is calculated using the predicted signed distance value \( f(x_i) \) and its gradient \( \nabla f(x_i) \):

\begin{align}
\label{eq:neural_pull}
x_{q} = x_i - f(x_i) \frac{\nabla f(x_i)}{\|\nabla f(x_i)\|}.
\end{align}

This operation facilitates effective gradient updates to the SDF by aligning \( x_q \) with the pseudo surface, which consists of target surface points \( x_t \) resampled from the multi-view rendering weights.

%
%
\subsection{Self-supervised Global Geometry Refinement}


\subsubsection{Self-supervision via multi-view rendering weights}
\label{sec: searching}
Recent research~\cite{bai2023dynamic} has demonstrated that rendering weights can effectively highlight regions of interest for sample allocation. As shown in \cref{fig:method}(a), these weights are automatically generated during the volume rendering process, they can serve as a form of self-supervision, providing valuable cues for localizing overlapping parts during training.
Moreover, NeuS~\cite{Wang_Liu_Liu_Theobalt_Komura_Wang_2021} demonstrates that, firstly, with an unbiased rendering weight function, surface points contribute more to their corresponding ray pixels than other ray samples;
Secondly, occlusion awareness ensures that the first intersection along a ray holds a higher value than subsequent intersections.
Consequently, the multi-view weights can be leveraged to generate a pseudo-surface for the global supervision.

As depicted in \cref{fig:method}(b), Hi-NeuS leverages this supervision by aggregating rendering weights from multiple views, enabling more frequent evaluations and thereby reducing uncertainty.
More importantly, differs from the ray-wise SDF constraint of NeuS in \cref{eq:neus_opacity}, our method imposes geometrical constraints on the 3D space, allowing reducing the neural SDF bias from a global scale.

To obtain this pseudo-surface, we utilize a temporary grid buffer to accumulate the multi-view rendering weights averaged across all camera poses.
We periodically refresh the buffer to update the training statistics.
Upon refreshing the grid buffer, we apply a global voting scheme, formulated as follows:
\begin{align}
\label{eq:vote}
w_t = \sum_r w_{i}\frac{1}{n_{i}}; \;  \ f(x_t) = \sum_r f(x_{i})\frac{1}{n_{i}};
\end{align}
where $n_{i}$ denotes the number of ray hits at the grid unit; $t$ denotes the buffer's refreshing times.
In addition, we also record signed distance values to facilitate subsequent refinement stages.
 %
 
Once $w_t$ is obtained, we apply a pulling operation to predict surface points $x_q$ that are expected to distribute around the pseudo-surface from $w_t$ when training converges.
To achieve it, we resample target surface points $x_t$ based on the recorded normalized rendering weights $w_{t-1}$. 
Our problem is then reformulated as the alignment of point clouds $\vec{x_q}$ and $\vec{x_t}$, subject to geometric constraints.
 
\begin{figure*}[t!]
    \centering
    \includegraphics[width=0.95\linewidth]{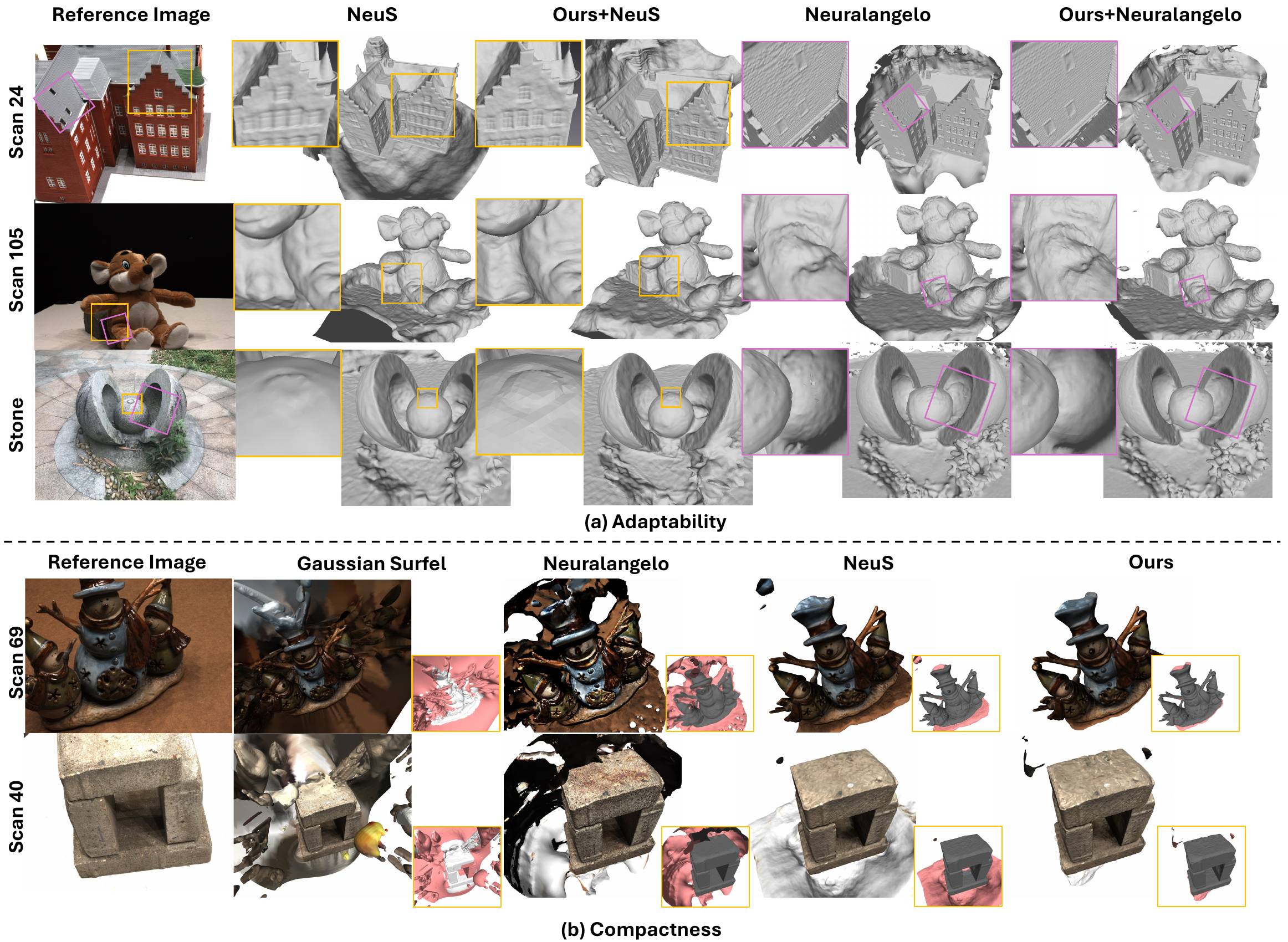}
    \caption{\textbf{Qualitative comparison of Hi-NeuS}: \textbf{(a) Adaptability.} We integrate our geometry refinement with NeuS and Neuralangelo to prove its adaptability. The magnified boxes reveal the recovered details.
    \textbf{(b) Compactness.} We compare our Hi-NeuS with the NeuS backbone against existing methods. The corner box of each image displays the highlighted areas outside the visual hull, denoted as \underline{mesh noise}. 
    }
    \label{fig:eval}
\end{figure*}
%
%

%
%
%
%

\subsubsection{Global Geometric Refinement} 
%
As illustrated in the bottom branch of \cref{fig:method}, we propose three geometric constraints to mitigate neural bias on a global scale: point cloud alignment loss $\mathcal{L}_{cd}$, global geometry consistency loss $\mathcal{L}_{global}$, and surface regularization loss $\mathcal{L}_{surf}$, which are introduced as follows:

\noindent\textbf{(1) Point cloud alignment loss ($\mathbf{CD\to 0}$}):
To align the predicted surface with $\vec{x}_t$, Hi-NeuS employs the bidirectional Chamfer Distance(CD). Specifically, CD is defined as $d(A, B) = \frac{1}{|\vec{x}_q|} \sum_{x \in A} \min_{x' \in B} \| x - x' \|_2^2$. 
This formula quantifies the similarity between two point sets, 
$A$ and $B$, by calculating the averaged nearest distance between each point in $A$ and any point in $B$. By minimizing the bidirectional CD, Hi-NeuS ensures both sets are matched as closely as possible in both directions. The point cloud alignment loss is formulated as:
\begin{align}
\label{eq:loss_cd}
\mathcal{L}_{cd}(\vec{x}_q, \vec{x}_t)&=\frac{1}{2}(d(\vec{x}_q, \vec{x}_t)+d(\vec{x}_t, \vec{x}_q)).
\end{align}
By minimizing $\mathcal{L}_{cd}(\vec{x}_q, \vec{x}_t)$,  
we guide the learnable surface points $\vec{x}_q$ to align toward $\vec{x}_t$, allowing for the refinement of the surface through the backpropagation of gradients into \(\vec{x}_q\) and subsequently into the SDF representation.

\noindent\textbf{(2) Global geometry consistency loss ($\mathbf{|SDF| < CD}$)}:
The SDF measures measures the orthogonal distance, representing the shortest distance from a given point to the surface boundary. However, during training, the learnable surface points
$\vec{x_q}$ may bring an inconsistent CD due to training dynamics and unstable learning. 
To maintain global geometry consistency, we require that the absolute SDF value of
 $\vec{x_q}$ remains consistently smaller than the CD of the point cloud. To enforce this constraint, we introduce a novel global geometry consistency loss, 
\begin{align}
\label{eq:loss_global}
\mathcal{L}_{global} = \big\vert|f(\vec{x}_q')| - d(\vec{x}_q', \vec{x}_t)\big\vert,
\end{align}
where $\vec{x}_{q}' \in \vec{x}_{q}$ is a filtered subset of points, ensuring that all predicted surface points lie within valid coordinate spaces.
Specifically, we derive $f(\vec{x}_{q}')$ from the grid buffer recorded at \cref{eq:vote}, excluding outliers that do not correspond to valid buffer units, to ensure stable training.

\noindent\textbf{(3) Surface regularization loss ($\mathbf{SDF \to 0}$)}:
Since $\vec{x}_q$ represents the surface points, the ideal signed distance values $f(\vec{x}_{q})$ should ideally be zero.
To reduce surface geometry bias, we introduce a penalty for the absolute SDF error on all valid learnable surface points $\vec{x}_q'$. This is captured by the surface SDF regularization loss $\mathcal{L}_{surf}$, defined as:
\begin{align}
\label{eq:surf_sdf}
\mathcal{L}_{surf} = f(\vec{x}_{q}') = \sum_i \frac{1}{n_{i}}|f(x'_{q, i})|.
\end{align}

\subsection{Optimization}
\label{sec:optimization}
To further verify the adaptability, we extend our framework to Neuralangelo~\cite{Neuralangelo}, which shares the same SDF-based rendering as NeuS~\cite{Wang_Liu_Liu_Theobalt_Komura_Wang_2021}.
In our implementation, we incorporate two additional losses: the eikonal loss $\mathcal{L}_{eik}$ and the curvature loss $\mathcal{L}_{curv}$.
The eikonal loss $\mathcal{L}_{eik}$ is based on the eikonal equation and ensures that the gradient magnitude is normalized throughout the entire space to reduce SDF truncation.
Meanwhile, the curvature loss $\mathcal{L}_{curv}$ guarantees that the analytical gradients of hash encoding are zero everywhere when using trilinear interpolation.
\begin{align}
\label{eq:loss_eik_curv}
\mathcal{L}_{eik} = \frac{1}{N} \sum_{i=1}^{N} \left( \left\| \nabla f(x_i) \right\|_2 - 1 \right)^2; \ \mathcal{L}_{\text{curv}} = \frac{1}{N} \sum_{i=1}^{N} \left| \nabla^2 f(x_i) \right|.
\end{align}
%
Our global geometric constraints $\mathcal{L}_{geo}$ are integrated as an add-on module, comprising a weighted sum of three components:\(\mathcal{L}_{cd}\), \(\mathcal{L}_{global}\), and \(\mathcal{L}_{surf}\):
\begin{align}
\label{eq:loss_total}
\mathcal{L}_{geo} &= \mathcal{L}_{cd} + w_{surf}\mathcal{L}_{surf}+ w_{global}\mathcal{L}_{global};
\end{align}
We then combine $\mathcal{L}_{geo}$ with other losses to form the total loss functions for NeuS and Neuralangelo:
\begin{align}
\mathcal{L}_{neus} &= \mathcal{L}_{RGB} + w_{eik}\mathcal{L}_{eik}+ w_{geo}\mathcal{L}_{geo}; \\
\mathcal{L}_{neuralangelo} &= \mathcal{L}_{neus} + w_{curv}\mathcal{L}_{\text{curv}}.
\end{align}
We apply these loss functions, $\mathcal{L}_{neuralangelo}$, $\mathcal{L}_{neus}$ in the original implementations for Neuralangelo and NeuS respectively.

\begin{table*}[t!]
  \centering
  \caption{Quantitative unmasked CD, PSNR, mesh noise results on DTU dataset~\cite{Jensen_Dahl_Vogiatzis_Tola_Aanaes_2014}. }
  \label{tab:dtu_all}
  \begin{adjustbox}{width=\linewidth,center}
    \begin{threeparttable}
    \begin{tabular}{llcccccccccccccccc} 
      \toprule[1.5pt]
\multirow{1}{*}{\rotatebox[origin=c]{90}{unmasked CD $\downarrow$}}
      & & 24 & 37 & 40 & 55 & 63 & 65 & 69 & 83 & 97 & 105 & 106 & 110 & 114 & 118 & 122 & Mean \\ 
      \cline{2-18}
      & Gaussian Surfels*\textdagger~\cite{dai2024highquality} &1.00 & 1.97 & 1.06 & 1.74 & 2.32 & 2.35 & 2.02 & 3.48 & 2.45 & 2.55 & 2.31 & 8.13 & 1.49 & 2.69 & 3.48 & 2.60\\
      \cline{2-18}
      &     NeuS~\cite{Wang_Liu_Liu_Theobalt_Komura_Wang_2021} & 1.59 & 1.98 & 1.44 & 0.95 & 1.82 & \cellcolor{orange!25}0.74 & \cellcolor{red!25}0.64 & 1.63 & \cellcolor{orange!25}1.30 & 1.41 & 0.59 & 1.33 & 0.44 & \cellcolor{orange!25}0.51 & \cellcolor{orange!25}0.54 & 1.13 \\
      & Neuralangelo*~\cite{Neuralangelo} &\cellcolor{orange!25}0.62 & 1.63 & \cellcolor{orange!25}0.66 & \cellcolor{orange!25}0.56 & 1.51 & 1.38 & 2.60 & 2.03 & 2.15 & 1.11 & \cellcolor{orange!25}0.46 & 1.31 & 0.48 & 0.95 & 1.25 & 1.25 \\    
      \cline{2-18}
      & Hi-NeuS(NeuS) & 0.96 & \cellcolor{red!25}0.93 & 0.71 & \cellcolor{red!25}0.47 & \cellcolor{red!25}1.37 & \cellcolor{red!25}0.71 & \cellcolor{orange!25}0.66 & \cellcolor{orange!25}1.45 & \cellcolor{red!25}1.02 & \cellcolor{red!25}1.07 & 0.58 & \cellcolor{orange!25}1.27 & \cellcolor{orange!25}0.44 & \cellcolor{red!25}0.50 & \cellcolor{red!25}0.54 & \cellcolor{red!25}0.81\\
      & Hi-NeuS(Neuralangelo)&\cellcolor{red!25}0.55 & \cellcolor{orange!25}1.55 & \cellcolor{red!25}0.61 & 0.60 & \cellcolor{orange!25}1.51 & 0.77 & 2.25 & \cellcolor{red!25}1.19 & 1.52 & \cellcolor{orange!25}1.09 & \cellcolor{red!25}0.43 & \cellcolor{red!25}1.20 & \cellcolor{red!25}0.43 & 0.88 & 1.36 & \cellcolor{orange!25}1.06 \\
      \midrule[1.2pt]
\multirow{9}{*}{\rotatebox[origin=c]{90}{unmasked CD $\downarrow$}}
      & RegSDF\textdagger~\cite{Zhang_Yao_Li_Fang_McKinnon_Tsin_Quan_2022} & 0.60 & 1.41 & 0.64 & 0.43 & 1.34 & 0.62 & 0.60 & \cellcolor{red!25}0.90 & \cellcolor{orange!25}0.92 & 1.02 & 0.60 & \cellcolor{red!25}0.59 & \cellcolor{red!25}0.30 & \cellcolor{orange!25}0.41 & \cellcolor{red!25}0.39 & 0.72 \\
      & NeuralWarp\textdagger~\cite{Darmon_Bascle_Devaux_Monasse_Aubry_2022} & 0.49 & \cellcolor{orange!25}0.71 & 0.38 & 0.38 & \cellcolor{red!25}0.79 & 0.81 & 0.82 & \cellcolor{orange!25}1.20 & 1.06 & \cellcolor{red!25}0.68 & 0.66 & 0.74 & 0.41 & 0.63 & 0.51 & \cellcolor{orange!25}0.68 \\
      & D-NeuS\textdagger~\cite{Chen_Zhang_Feldmann_Schreer_Eisert_2022} &0.44 & 0.79 &\cellcolor{orange!25}0.35 & \cellcolor{red!25}0.39 &0.88 &\cellcolor{orange!25}0.58 &\cellcolor{red!25}0.55 &1.35 & \cellcolor{red!25}0.91 &0.76 & \cellcolor{red!25}0.40 & \cellcolor{orange!25}0.72 & \cellcolor{orange!25}0.31  &\cellcolor{red!25}0.39 & \cellcolor{orange!25}0.39 & \cellcolor{red!25}0.61\\
      \cline{2-18}
     & NeRF~\cite{Mildenhall_Srinivasan_Tancik_Barron_Ramamoorthi_Ng_2020} & 1.90 & 1.60 & 1.85 & 0.58 & 2.28 & 1.27 & 1.47 & 1.67 & 2.05 & 1.07 & 0.88 & 2.53 & 1.06 & 1.15 & 0.96 & 1.49 \\
      & VolSDF~\cite{Yariv_Gu_Kasten_Lipman_2021} & 1.14 & 1.26 & 0.81 & 0.49 & 1.25 & \cellcolor{orange!25}0.70 & 0.72 & 1.29 & 1.18 & 0.70 & 0.66 & 1.08 & 0.42 & 0.61 & 0.55 & 0.86 \\
      & NeuS~\cite{Wang_Liu_Liu_Theobalt_Komura_Wang_2021} &0.93 &1.07 &0.81 &0.38 &1.02 &0.60 &\cellcolor{orange!25}0.58 &1.42 &1.15 &0.78 &0.57 &1.16 &0.35 &0.45 &0.46 &0.78 \\
      & Neuralangelo*~\cite{Neuralangelo} & \cellcolor{orange!25}0.39 & 0.72 & \cellcolor{red!25}0.35 & \cellcolor{orange!25}0.33 & \cellcolor{orange!25}0.82 & 0.74 & 1.70 & 1.34 & 1.95 & 0.71 & 0.47 & 1.00 & 0.33 & 0.82 & 0.78 & 0.83 \\ 
      \cline{2-18}
    & Hi-NeuS(NeuS)  & 0.77 & 0.90 & 0.73 & 0.37 & 1.00 & 0.59 & 0.59 & 1.42 & 1.19 & 0.79 & 0.56 & 1.93 & 0.35 & 0.45 & 0.48 & 0.81\\
      & Hi-NeuS(Neuralangelo) &\cellcolor{red!25}0.39 &\cellcolor{red!25}0.71 &0.36 &\cellcolor{red!25}0.33 &0.92 &\cellcolor{red!25}0.55 &1.42 &1.25 &1.44 &0.73 &\cellcolor{orange!25}0.45 &0.99 &0.33 &0.70 &0.73 &0.75 \\
      \midrule[1.2pt]
      \multirow{7}{*}{\rotatebox[origin=c]{90}{PSNR $\uparrow$}} 
      & 
RegSDF\textdagger~\cite{Zhang_Yao_Li_Fang_McKinnon_Tsin_Quan_2022} &24.78 &25.31 &23.47 &23.06 &22.21 &28.57 &25.53 &21.81 &28.89 &26.81 &27.91 &24.71 &25.13 &26.84 &21.67 &28.25  \\
 \cline{2-18}
      & VolSDF~\cite{Yariv_Gu_Kasten_Lipman_2021} & 26.28 & 25.61 & 26.55 & 26.76 & 31.57 &31.50 &29.38& 33.23 &28.03 &32.13& 33.16& 31.49& 30.33& 34.90& 34.75& 30.38  \\
      & NeRF~\cite{Mildenhall_Srinivasan_Tancik_Barron_Ramamoorthi_Ng_2020} & 26.24 & 25.74 & 26.79 & 27.57 & 31.96 & 31.50 & 29.58 & 32.78 &28.35 & 32.08 & 33.49 & 31.54 & 31.00 & 35.59 & 35.51 & 30.65 \\
      & NeuS~\cite{Wang_Liu_Liu_Theobalt_Komura_Wang_2021} &25.82 &23.64 &26.64 &25.60 &27.68 &30.83 &27.68 &34.04 &26.61 &31.35 &29.29 &28.08 &28.55 & 31.28 &33.68 &28.79 \\
            & Neuralangelo*~\cite{Neuralangelo} & \cellcolor{red!25}30.90 & \cellcolor{orange!25}28.01 & \cellcolor{red!25}31.60 & \cellcolor{red!25}34.18 & \cellcolor{red!25}36.15 & \cellcolor{red!25}36.30 & \cellcolor{red!25}34.10 & \cellcolor{orange!25}38.84 & \cellcolor{red!25}31.28 & \cellcolor{orange!25}37.15 & \cellcolor{red!25}35.73 & \cellcolor{orange!25}33.60 &\cellcolor{red!25}31.80 & \cellcolor{red!25}38.19 & \cellcolor{red!25}38.42 & \cellcolor{red!25}34.13 \\ \cline{2-18}   
      & Hi-NeuS(NeuS) & 26.24 & 23.79 & 26.98 & 25.70 & 30.21 & 31.65 & 29.27 & 34.94 & 26.59 & 32.31 & 32.37 & 29.30 & 28.73 & 34.15 & 33.69 & 29.73 \\
      & Hi-NeuS(Neuralangelo) &\cellcolor{orange!25}30.80 &\cellcolor{red!25}28.01 &\cellcolor{orange!25}31.50 &\cellcolor{orange!25}29.82 &\cellcolor{orange!25}36.12 &\cellcolor{orange!25}36.17 &\cellcolor{orange!25}34.06 & \cellcolor{red!25}39.04 &\cellcolor{orange!25}31.13 &\cellcolor{red!25}37.18 &\cellcolor{orange!25}35.62 &\cellcolor{red!25}33.71 &\cellcolor{orange!25}31.53 &\cellcolor{orange!25}38.01 &\cellcolor{orange!25}38.07 &\cellcolor{orange!25}34.05 \\
      \midrule[1.2pt]
      \multirow{5}{*}{\rotatebox[origin=c]{90}{Noise\% $\downarrow$}} 
                        & Gaussian Surfels*\textdagger~\cite{dai2024highquality} & 43.09 & 45.46 & 50.04 & 61.64 & \cellcolor{red!25}25.07 & 60.11 & 58.98 & \cellcolor{orange!25}62.56 & 54.89 & 56.93 & 75.41 & 99.69 & 77.05 & 74.77 & 84.89 & 62.04 \\
                        \cline{2-18} & NeuS~\cite{Mildenhall_Srinivasan_Tancik_Barron_Ramamoorthi_Ng_2020}&40.75 &60.50 &56.83 &72.60 &32.27 &\cellcolor{red!25}28.69 &\cellcolor{orange!25}26.07 &75.41 &43.14 &64.46 &57.33 &\cellcolor{red!25}17.35 &\cellcolor{orange!25}15.47 &\cellcolor{red!25}8.53 &\cellcolor{red!25}11.03 &\cellcolor{orange!25}39.13 \\
        & Neuralangelo*~\cite{Neuralangelo}&36.24 &52.32 &55.62 &66.63 &56.77 &57.84 &77.97 &76.70 &57.71 &63.60 &39.52 &84.71 &49.13 &35.34 &51.41 & 57.44  \\
      \cline{2-18}
         & Hi-NeuS(NeuS)& \cellcolor{orange!25}34.02 & \cellcolor{red!25}3.74 & \cellcolor{red!25}5.90 & \cellcolor{red!25}49.12 & \cellcolor{orange!25}27.58 & \cellcolor{orange!25}29.52 & \cellcolor{red!25}22.04 & 67.33 & \cellcolor{red!25}26.98 & 61.79 & \cellcolor{red!25}32.90 & 19.47 & \cellcolor{red!25}14.71 & \cellcolor{orange!25}17.10 & \cellcolor{orange!25}15.33 & \cellcolor{red!25}28.50\\
      & Hi-NeuS(Neuralangelo) &\cellcolor{red!25}32.36 &\cellcolor{orange!25}44.25 &\cellcolor{orange!25}39.16 &\cellcolor{orange!25}59.96 &43.01 &34.23 &68.31 &\cellcolor{red!25}61.89 &58.68 &\cellcolor{red!25}60.71 &\cellcolor{orange!25}36.15 &\cellcolor{orange!25}17.45 &21.54 &28.42 &57.60 &45.67 \\
      \bottomrule[1.5pt]
    \end{tabular}%
    \begin{tablenotes}
\item * 
%
\textdagger \ denotes auxiliary data inputs, including 3D points from SFM or other pretrained models.
* denotes our evaluated results with the available open-source configuration. The best performance is highlighted in \textcolor{red}{red}, while the second best is marked in \textcolor{orange}{orange} for each measure and scene. Hi-NeuS(backbone) refers to the backbone architecture used in conjunction with our proposed geometric constraints.
\end{tablenotes}
\end{threeparttable}
\end{adjustbox}
\end{table*}

%


\section{Experiments}
\noindent\textbf{Datasets.}
We conducted experiments on the DTU dataset~\cite{Jensen_Dahl_Vogiatzis_Tola_Aanaes_2014}, which includes 15 object-centric scenes. Each scene comprises 49 or 64 images captured by a robot-held monocular RGB camera, with ground truth obtained using a structured light scanner. Additionally, following NeuS, we performed experiments on 7 challenging scenes from the low-resolution subset of the BlendedMVS dataset~\cite{Yao_Luo_Li_Zhang_Ren_Zhou_Fang_Quan_2020}. Each of these scenes contains 31 to 143 images at $768 \times 576$ pixels with corresponding masks. 
Moreover, we capture hand-held phone videos using an iPhone 13, recording short sequences of 30 to 60 seconds in duration, with 80-160 frames sampled uniformly throughout each sequence.
All mesh reconstructions in this study were processed using the marching cubes algorithm, with the resolution set to 512.
\begin{figure}[t!]
    \centering
    \includegraphics[width=\linewidth]{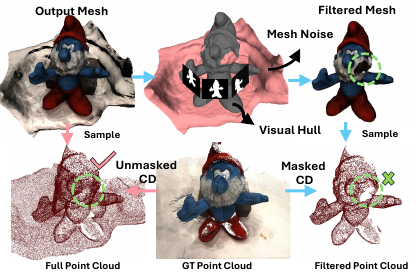}
    \caption{\textbf{The mesh post-processing and its evaluation}: Mesh noise refers to the space ratio outside the 3D visual hull created by silhouettes. The dashed circles highlight the areas where space is missing and must be evaluated. To evaluate this, we use sampled point clouds and GT point clouds to calculate the CD between the two. We compare the range of space to be evaluated between our proposed unmasked CD (\textcolor{red}{red arrows}) and the masked CD used in previous methods~\cite{Neuralangelo, Wang_Liu_Liu_Theobalt_Komura_Wang_2021, Oechsle_Peng_Geiger_2021, Yariv_Kasten_Moran_Galun_Atzmon_Basri_Lipman_2020}  (\textcolor{blue}{blue arrows}).  
    }
    \label{fig:eval_pipe}
\end{figure}

\noindent\textbf{Evaluation criteria.} 
We calculate the Peak Signal-to-Noise Ratio (PSNR) on all masked parts of images from each scene, following the same method as Neuralangelo.
We report the masked CD on the observed regions with the masked-out meshes to ensure fair comparison against previous works~\cite{Neuralangelo, Wang_Liu_Liu_Theobalt_Komura_Wang_2021, Darmon_Bascle_Devaux_Monasse_Aubry_2022, Zhang_Yao_Li_Fang_McKinnon_Tsin_Quan_2022, Chen_Zhang_Feldmann_Schreer_Eisert_2022, Yariv_Gu_Kasten_Lipman_2021, Mildenhall_Srinivasan_Tancik_Barron_Ramamoorthi_Ng_2020}. 
We propose evaluating meshes using the unmasked CD, which assesses the extracted raw mesh without any post-processing.
As illustrated in \cref{fig:eval_pipe}, our empirical study reveals that due to the limited number of camera views, the visual hull may inadvertently cover desirable object parts, resulting in an incomplete filtered mesh. To address this issue, we consider using the raw mesh directly for a more comprehensive and error-free evaluation of meshes on a global scale.
For evaluating geometry noise, we introduce a mesh noise metric that calculates the ratio of filtered faces to the total number of mesh faces during mesh filtering as depicted in \cref{fig:eval_pipe}. This metric effectively indicates the majority of regions that users do not intend to have uncovered by the multi-view object masks.

\noindent\textbf{Baselines.} The baselines are as follows:
\textbf{1)} 
NeuS~\cite{Wang_Liu_Liu_Theobalt_Komura_Wang_2021}, a pioneering work that first developed SDF-based volume rendering, has had a profound impact on the surface reconstruction field. It has inspired subsequent research, including Neuralangelo.  The NeuS's superior performance justifies its exclusion from direct comparison with contemporaries like UNISURF\cite{Oechsle_Peng_Geiger_2021} and IDR~\cite{Yariv_Kasten_Moran_Galun_Atzmon_Basri_Lipman_2020} in \cref{tab:dtu_all}. However, it is worth noting that NeuS's SDF representation introduces a significant amount of noise, approximately 40\%, which underscores our argument that geometry is still underconstrained when relying solely on ray-based geometry constraints.
\textbf{2)} 
Neuralangelo~\cite{Neuralangelo} is a state-of-the-art surface reconstruction framework that can recover high-quality surfaces. However, due to the lack of publicly available per-scene configurations, we apply the same configuration to all DTU scenes, without performing per-scene fine-tuning.
To ensure a fair comparison, our Hi-NeuS framework, which builds upon Neuralangelo, also utilizes this same configuration.
 Unfortunately, Neuralangelo introduces considerable noise, with approximately 60\%. These artifacts necessitate the use of foreground masks for cleanup in DTU datasets, which can be problematic as visual hulls alone may not accurately capture the intended geometry. (See \cref{fig:eval_pipe} for an illustration of this issue.)

\noindent\textbf{Implementation details.}
We adopt the same experimental setup and framework architectures as NeuS and Neuralangelo. For detailed hyperparameter configurations, we refer readers to the original implementations of NeuS and Neuralangelo.
%
In our method, we refine learnable surface points $\vec{x}_q$ to a subset $\vec{x}_q’$, comprising only the valid points within the normalized coordinate space $[-1, 1]$. The resampled point cloud supervision, $\vec{x}_t$, matches the set size of $\vec{x}_q$,  
as described in \cref{eq:loss_cd}. This enables global surface refinement to have equal strength for both directions between the two point clouds.
%
%
We utilize a grid buffer with a resolution of $(32, 64, 128)$ implemented using ~\cite{Yu_Li_Tancik_Li_Ng_Kanazawa_2021}, which benefits from customized CUDA acceleration for efficient spatial queries. The grid buffer is reset to 0 after considering a certain ratio of views, allowing it to record the most recent training statistics after refreshing.

\subsection{Performances}
\noindent{\textbf{Quantitative \& Qualitative Results.}}
Hi-NeuS achieves comparable or superior performance in terms of both masked and CD for most scenes in DTU. Specifically, as shown in \cref{tab:dtu_all} and \cref{fig:eval}, the CD of scene 24 is reduced by approximately 17\%, allowing for more detailed architectural features to be captured on the building.
Especially,
as shown in (b) of \cref{fig:scale_loss}, Hi-NeuS successfully recovers the missing hand of the clock, which NeuS fails to reconstruct, likely due to a lack of front-facing photos. This demonstrates Hi-NeuS's ability to effectively capture the 3D structure and depict the correct geometry when views are limited, proving the effectiveness of the global scale geometric constraint.
%
Besides, our method recovers more intricate details in scenes such as 105, where the brick structures behind the toy are more accurately represented. However, the masked CD calculation does not fully reflect this improvement, as it is limited by the visual hull coverage. In contrast, our proposed unmasked CD provides a more comprehensive evaluation of the raw mesh geometry, revealing the overall geometry accuracy.
%
In terms of visual quality, Hi-NeuS demonstrates improved results on the NeuS backbone and achieves comparable PSNR values on Neuralangelo. This suggests that the corrected geometry may have a positive impact on color learning, potentially leading to more accurate and realistic color representations. 
Regarding noise reduction, Hi-NeuS achieves superior performance, reducing noise by approximately 37\% compared to the NeuS backbone and by around 20\% compared to Neuralangelo. This improvement is visually evident, with a significant reduction in noise scale.
In general, Hi-NeuS demonstrates its ability to recover compact, accurate, and high-fidelity surfaces, showcasing its adaptability and versatility when integrated with NeuS backbones.
Our study's full quantitative and qualitative result is attached at Tab.1 in \textit{suppl.}.

%


\subsection{Ablation Studies and Analysis}
\begin{figure}[t!]
    \centering
    \includegraphics[width=\linewidth]{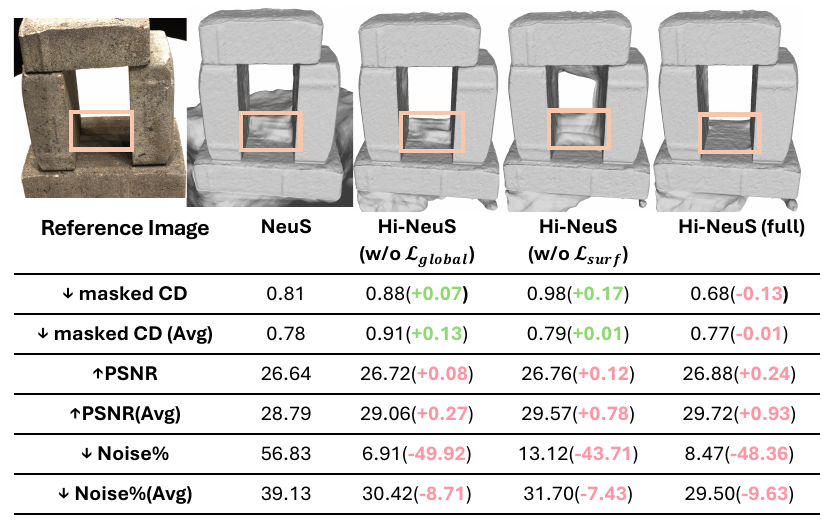}
    \caption{\textbf{Ablation study on proposed losses:} performance evaluation on scan 40 in the DTU dataset and the average results across the DTU dataset, with their performance comparisons relative to NeuS. The boxes emphasize the difference in mesh quality. }
    \label{fig:abls_40}
\end{figure}

\noindent\textbf{Loss Effectiveness.}
To gain insight into the impact of our proposed geometrical constraints on reconstruction results, we evaluate the individual loss components in \cref{eq:loss_total} and present the results in \cref{fig:abls_40} for the challenging Scene 40 from the DTU dataset. The highlighted area reveals the differences in texture inside the opening.
%
Our analysis shows that the brick structure with an opening in the center is reconstructed with high fidelity, capturing its fine geometric details. Notably, the global geometry consistency loss $\mathcal{L}_{global}$ effectively aligns surface points to more accurate positions, resulting in a 15\% improvement in masked CD. However, this improvement comes at the cost of increased noise, with increased 22\% noise observed after adding $\mathcal{L}_{global}$, as exemplified by artifacts at the bottom of the brick structure.
%
In contrast, the surface regularization loss $\mathcal{L}_{surf}$ efficiently captures the surface boundary by penalizing absolute SDF errors on its zero-level set. This leads to a 6\% reduction in noise and a marginal improvement in the CD measure.
Combining both $\mathcal{L}_{global}$ and $\mathcal{L}_{surf}$, our full model achieves a more detailed reconstruction while maintaining compactness. Notably, the rendering fidelity is significantly enhanced by reducing geometric errors for both loss components, demonstrating the effectiveness of our proposed geometrical constraints.

\begin{figure}[t!]
    \centering
    \includegraphics[width=\linewidth]{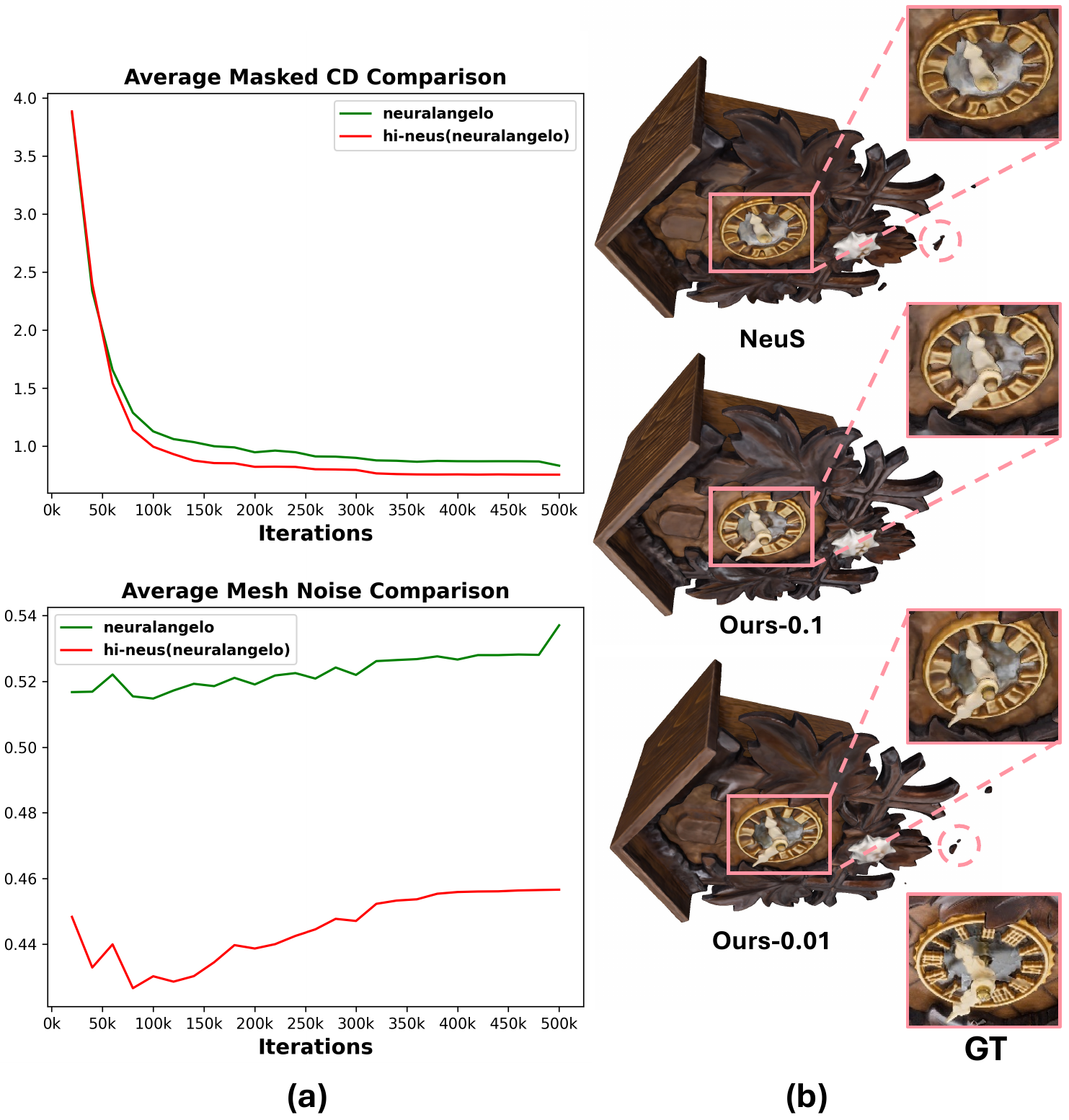}
    \caption{\textbf{(a)} Average masked CD and mesh noise on Neuralangelo and our revised Neuralangelo across DTU datasets during training. \textbf{(b)} Effectiveness of loss scale refinement on the BlendedMVS dataset with varying $\mathcal{L}_{geo}$
  scales. The magnified box provides a detailed comparison with GT, while the dashed circles highlight the reduction of artifacts.
    }
    \label{fig:scale_loss}
\end{figure}
\begin{figure}[t!]
    \centering
    \includegraphics[width=\linewidth]{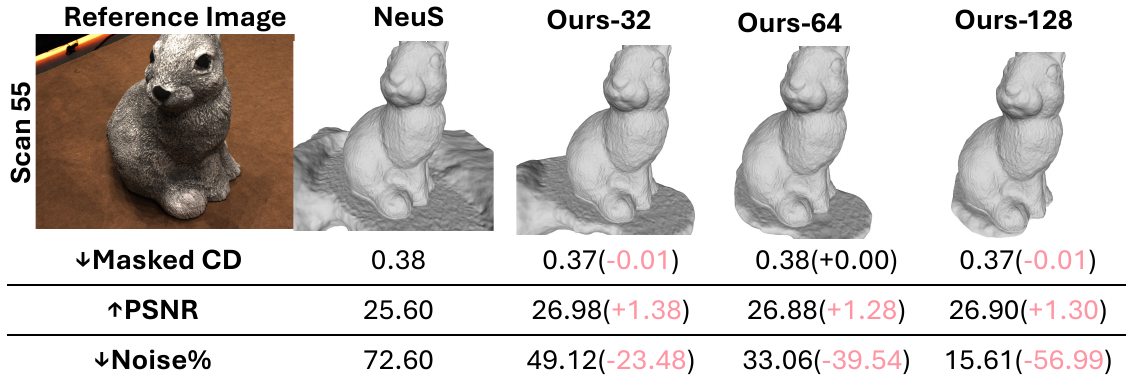}
    \caption{\textbf{The effectiveness of grid resolution}: refinement on DTU scan 55 by varying grid resolutions.
    }
    \label{fig:scale_res}
\end{figure}
\begin{figure}[t!]
    \centering
    \includegraphics[width=\linewidth]{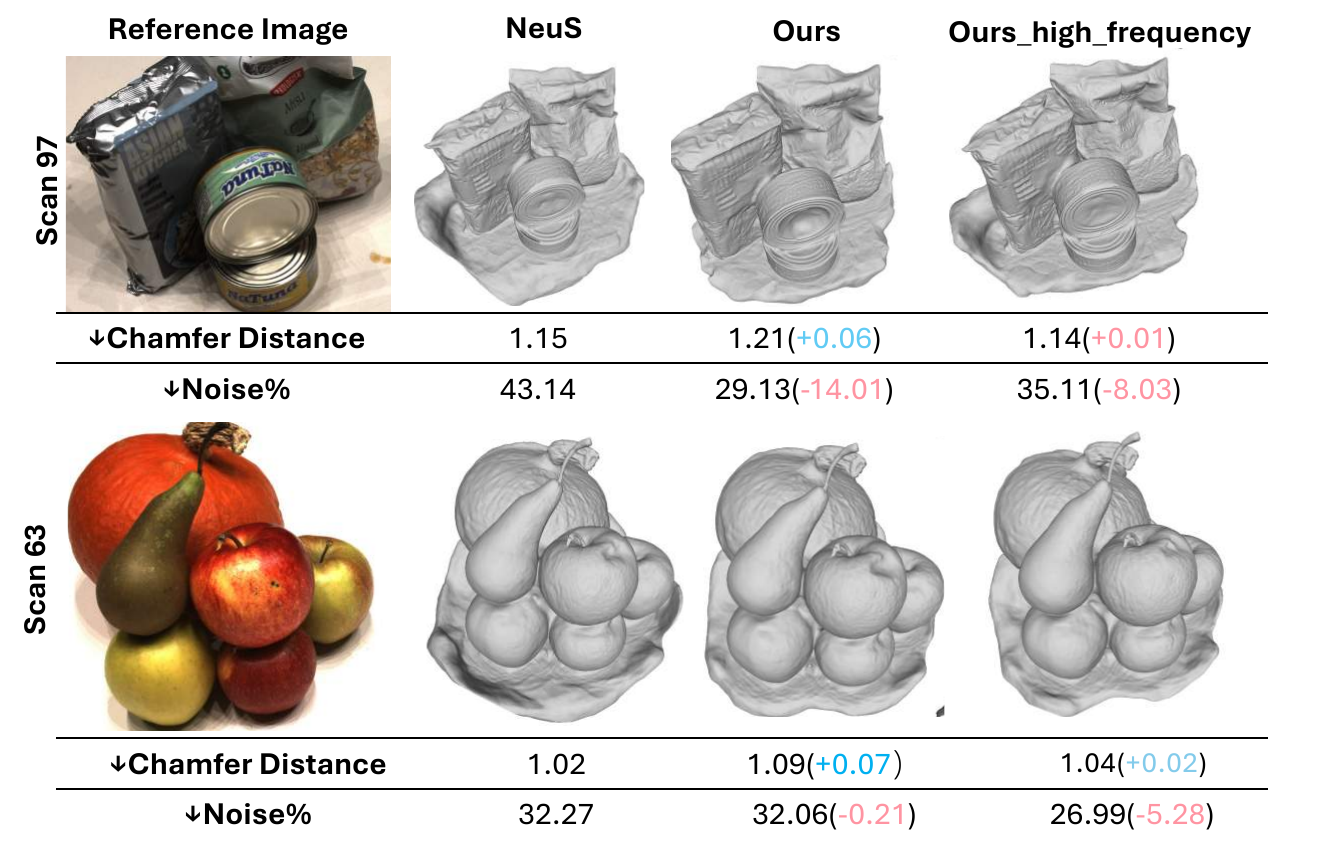}
    \caption{\textbf{To solution on the highly reflective case}: We increase our proposed gird buffer refreshing frequency and compare the results with the previous version.
    }
    \label{fig:abl_freq}
\end{figure}
\noindent\textbf{Analysis on the training process. } As shown in (a) of \cref{fig:scale_loss}, we averaged each iteration across all scenes in the DTU dataset using the Neuralangelo backbone and compared it with our Hi-NeuS model. Our findings indicate that the masked CD for Hi-NeuS is consistently lower than that of the baseline, especially during the early training stages between 50k and 200k iterations. Additionally, our model exhibits significantly less noise throughout the training process, with the noise level reaching its minimum around 100k iterations and then gradually increasing. In contrast, Neuralangelo experiences a sharp noise increase, especially at the end of training. These observations demonstrate that our model is more accurate and stable and converges faster compared with the baseline.
We refer interested readers to Sec.4.1 of our \textit{suppl.} for more analysis and visual demonstrations.

\noindent\textbf{Strength of Geometric Refinement. }
We investigate the effectiveness of Hi-NeuS in adjusting the strength of geometric refinement through the grid buffer resolution and loss scale.
Increasing the grid resolution enhances the accuracy of the grid in recording more fine-grained rendering weights, leading to a more compact structure. As shown in \cref{fig:scale_res}, scene 55 achieves a more compact result without compromising PSNR or noise. However, it is worth noting that increasing the resolution may not be universally beneficial, particularly for scenes with highly reflective materials, where it may lead to uneven distribution of rendering weights.
As shown in \cref{fig:abl_freq}, the metallic material in scene 97 exhibits CD degradation, as depicted in the distorted tin surface. Specifically, the high reflectivity in one direction causes the SDF to blend toward the reflected directions, resulting in a less noisy but more distorted surface.
On the other hand, as depicted in (b) of \cref{fig:scale_loss}, adjusting the loss scale has a similar effect on the compactness and accuracy of the mesh output. 
Increasing the loss scale produces more compact mesh outputs with reduced noise. In comparison to a loss scale of 
0.01,
a higher loss scale like 0.1 may result in less noise given more intensive constraints.
Overall, our results indicate that Hi-NeuS effectively scales the global refinement with varying grid resolutions and loss scales, ensuring optimal performance in different scenarios.
%

\noindent\textbf{Our solution for highly reflective scenarios. } As previously discussed, our method can be challenging to handle in scenarios with high reflectance variations, such as metals in scene 97 and smooth fruit surfaces in scene 63. To address this, we introduce a hyperparameter to regulate the buffer update frequency. Instead of collecting images from all camera views, we refresh the buffer from scratch after a certain number of views to record upcoming values.
This approach is particularly effective in scenarios with challenging light reflections, as illustrated in \cref{fig:abl_freq}. For example, Scan 97 in the DTU dataset features highly reflective surfaces, resulting in intensive light contributions from accumulated perspectives. This leads to distortion in the mesh despite lower noise levels.
To mitigate this issue, we employ a higher frequency update, using fewer images to accumulate rendering weights, which provides more instant feedback on the ongoing training status. As shown in \cref{fig:abl_freq}, the higher frequency update substantially reduces distortion.
This approach strikes a balance between maintaining consistency across multiple views and mitigating distortion introduced by buffer delay. 
%

\noindent\textbf{Training time and potential improvements. } Our proposed global geometry refinement module requires additional time compared to the NeuS backbone. On average, training time across scenes is approximately 30\% slower. However, testing time remains unchanged, as the SDF and color fields have the same parameter size and we use the same marching cube resolution.
In the NeuS's setup, our training time for one case was around 10 hours, compared to the 8-hour baseline.
To further accelerate \textit{real-world} applications, we integrate our adaptable geometrical constraints into the Instant-NGP implementation with improved CUDA parallelism~\cite{instant-nsr-pl}.
This integration significantly reduces training time, from \textit{around 10 minutes} per case for our adapted algorithms in the new module.
\begin{figure}[t!]
    \centering
    \includegraphics[width=\linewidth]{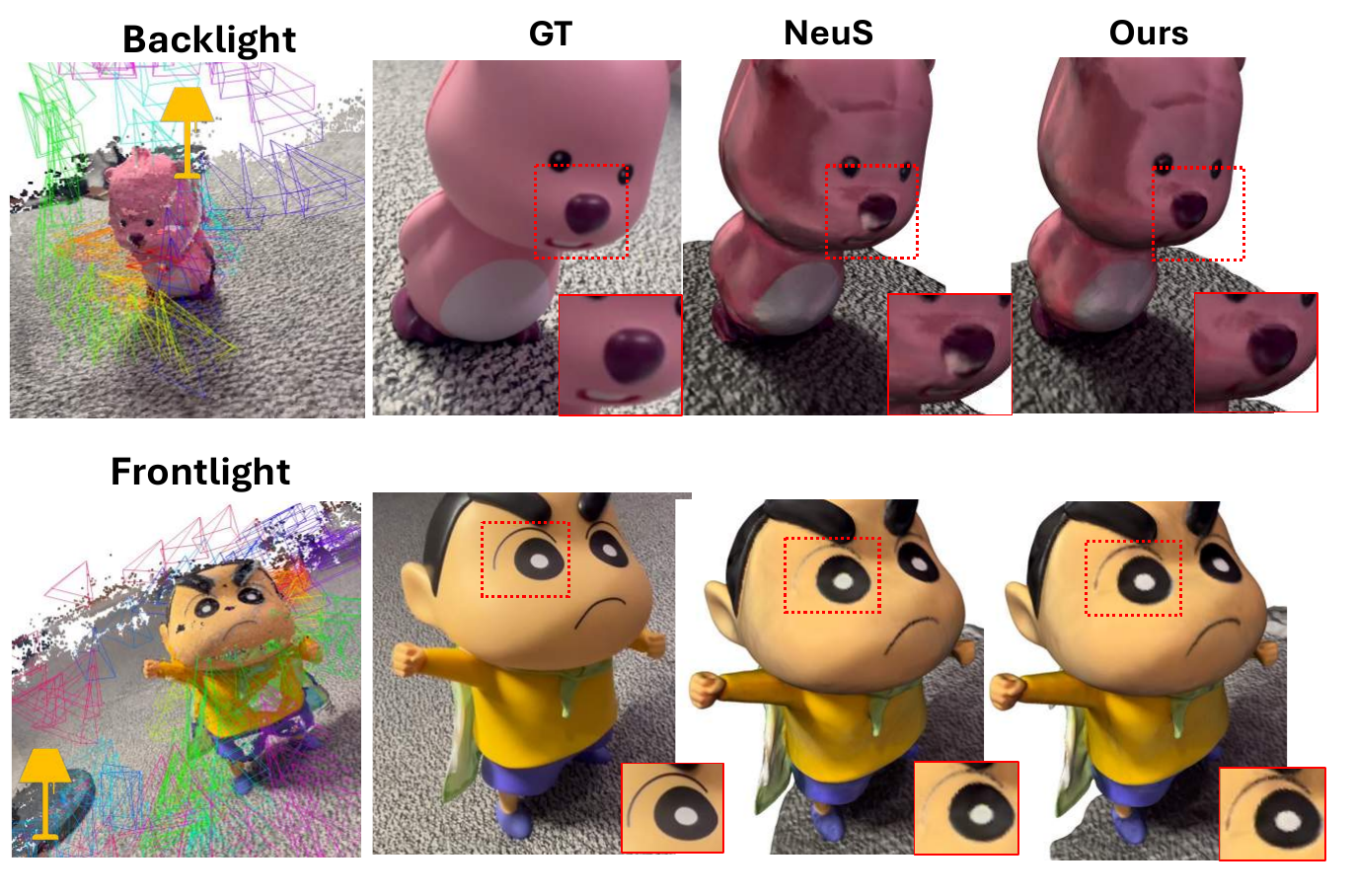}
    \vspace{-16pt}
    \caption{\textbf{Real-world application with our handheld phone captures}: We compare different lighting conditions, including back and front lights. The yellow icon denotes where our laptop is set.
The magnified boxes reveal the visual quality differences.
    }
    \label{fig:hand_eval}
\end{figure}

\subsection{Phone Capturing and Reconstruction Pipeline}
We present our pipeline for capturing and reconstructing objects from phone-captured videos to 3D meshes. The pipeline consists of three stages: (1) preprocessing videos with COLMAP to estimate camera poses, (2) training Hi-NeuS to learn the underlying geometry, (3) extracting meshes from the learned SDF and color fields, and (4) editing meshes on software for artistic creation.
In this study, we focus on object-centric capturing. For more challenging real-world scenarios, such as forward-facing and aerial circling, please refer to Section 5 of our \textit{suppl.} material.

\noindent\textbf{(1) Preprocessing videos with COLMAP. }
Given a short video, we use COLMAP\cite{Schnberger2016StructurefromMotionR} to estimate camera poses. Before running COLMAP, we evenly sample the video frames to around 80 or 160 frames. As shown in \cref{fig:hand_eval}, we visualize estimate camera trajectories and the dense map output of COLMAP. Note that in practice, only the sparse mapping is required for pose estimation. The COLMAP processing takes approximately 2 minutes using the exhaustive matching method for optimal pose quality.

\noindent\textbf{(2) Launch Hi-NeuS training. }
With posed RGB frames, we perform geometry refinement using NeuS or its variants. As illustrated in \cref{fig:method}, we first gather multi-view rendering weights and set a grid buffer to manage them during training. Our global geometry refinement process then reduces geometry bias during training.

\noindent\textbf{(3) Extract meshes. }
Given a batch of grid samples and camera poses, we evaluate each sample's signed distance values and vertex colors. We then perform marching cubes on the signed distance values to extract the surface and assign each surface vertex with its corresponding color.

\begin{figure}[t!]
    \centering
    \includegraphics[width=\linewidth]{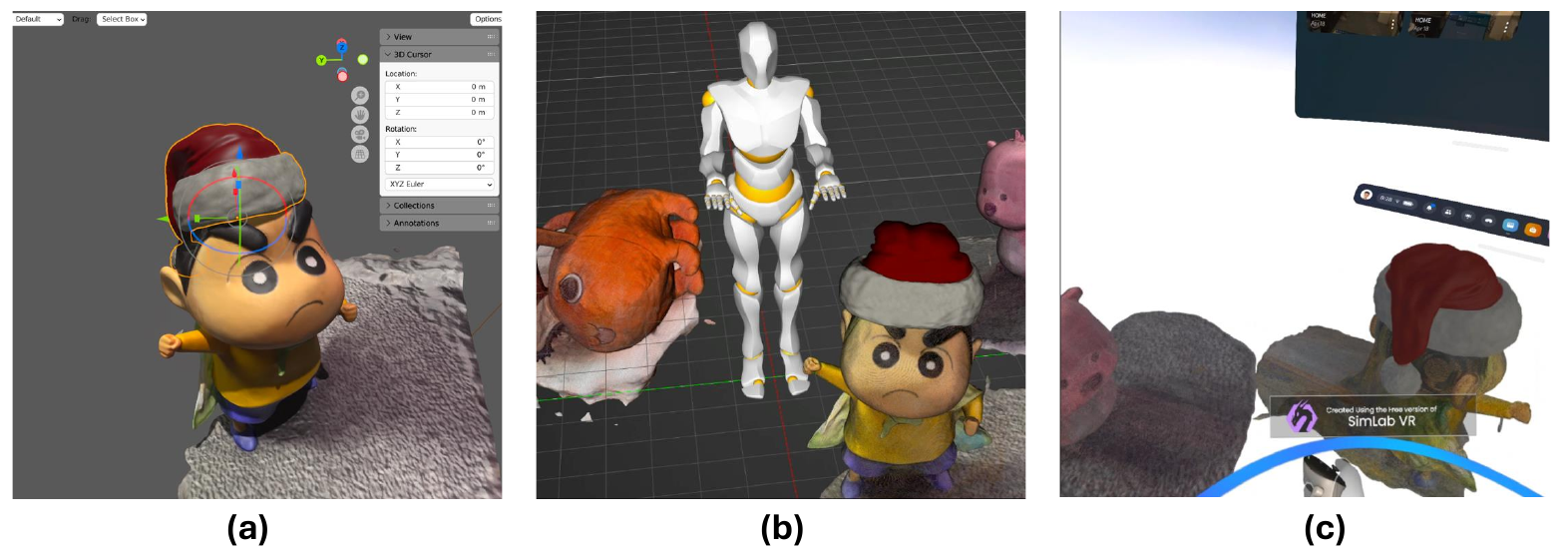}
    \vspace{-16pt}
    \caption{\textbf{Cross-platform viewing and editing}: \textbf{(a)} The mesh editing in Blender. 
    \textbf{(b)} The mesh composition by SimLab Composer. 
    \textbf{(c)} The mesh viewing with VR headset. 
    }
    \label{fig:vr}
\end{figure}

\noindent\textbf{(4) Content creation in computer platform. }
Most extracted meshes can be used directly or with minor edits, thanks to our proposed algorithm. As shown at \cref{fig:vr} (a), we further refine them using software such as Meshlab\cite{Cignoni2008MeshLabAO} and Blender\cite{blender} for mesh editing. For instance, as shown in \cref{fig:teaser}, we added a Santa hat to the toy's head and adjusted the lighting in the scene to enhance coherence during rendering. 

\noindent\textbf{(5) Application in VR. }
The reconstructed objects can then be combined in SimLab Composer~\cite{SimLabSoft} from the computer platform like \cref{fig:vr} (b) and put into the headset for an immersive VR/AR experience like \cref{fig:vr} (c).
The pipeline streamlines the surface reconstruction from object-centric views.

\noindent\textbf{Discussion on the light condition. }
In our phone-capturing process, we find that lighting conditions are a critical factor in obtaining high-quality meshes. Specifically, when comparing backlight and frontlight conditions in ~\cref{fig:hand_eval}, we observe that backlight conditions often result in dark artifacts or shallow details, whereas frontlight conditions tend to capture more realistic details.
To achieve optimal results, users are recommended to capture objects under well-lit conditions, ideally with the light source positioned in front of the object.
Furthermore, in terms of quality validation, our geometric refinement approach provides more realistic textures compared to NeuS, thanks to the refined geometry.

\section*{Supplemental Materials}
\label{sec:supplemental_materials}
We direct readers to our \textit{suppl.} materials for the video demonstration, which includes the complete processing pipeline and visual comparisons with rotating camera views. Additionally, the paper's other discussions are also included.

\section{Conclusion and Future work}
In this paper, we introduced Hi-NeuS, a novel rendering-based neural implicit surface reconstruction framework that leverages SDF-based volume rendering with our proposed global geometrical constraint. Our algorithm enabled recovering more compact and precise surfaces without relying on multi-view object masks.
 The capability and performance of our framework have been rigorously tested against the SOTA models with various datasets, demonstrating superior generalized performance in reducing geometry errors and recovering intricate details. 
By streamlining the geometry-capturing process, our framework has the potential to enable the geometry extraction directly from phone-captured data to meshes. This reduces the need to annotate multi-view object masks, facilitating seamless viewing and content creation in VR/AR.

\noindent\textbf{Future work.} We plan to adapt Hi-NeuS to more baselines and other datasets to further verify its ability. Furthermore, we would like to explore and execute more effective geometry constraints to boost our geometry accuracy. 









\acknowledgments{
The authors wish to thank A, B, and C. This work was supported in part by
a grant from XYZ.}

\bibliographystyle{abbrv-doi}

\bibliography{template}

\clearpage






\maketitle

In this supplementary document, we provide (1) The algorithm overview along with other implementation details; (2) further discussion on the multi-object annotation; (3) additional ablation studies, different types of real-world capturing, and full quantitative and qualitative results; (4) the video demos.

\section{Additional Implementation details}
In the global surface searching module illustrated in Sec.3.3.1, , we add contrast on the collected rendering weights based on buffer statistics values,
\(w_i' = max(w_i + \delta(w_i-\frac{1}{n}\sum_i w_i), 0)\), where $w_i$ is volume rendering weight, and $\delta$ is a hyperparameter to balance the strength of the contrast adjustment. 
This method can reduce the ambient noise while making it more possible to sample on more valuable surface regions. 
\begin{algorithm}[h]
\caption{Iteration $t$ of Hi-NeuS Training}\label{alg:cap}
\textbf{Input}: camera poses, RGB pixels, grid buffer; 
\textbf{Output}: predicted pixel color $\hat{c}_{r_i}$, global geometric constraints $\mathcal{L}_{geo}$.
\begin{algorithmic}[1]
\IF{\textbf{Iter} t $>$ 1}
    \STATE Access buffer for weights $\vec{w_{t-1}}$ and SDF $f(\vec{x_{t-1}})$.
    \STATE Resample supervision $\vec{x_{t}} \sim P(x_{t-1}|w_{t-1})$
\ENDIF
\FOR{\textbf{pose} $=0$ \textbf{to} \textbf{MaxPose}}
    \STATE Sample $H \times W$ rays from a the given camera pose.
    \FOR{$i=0$ \textbf{to} $H \times W$}
        \STATE Calculate ray-grid intersection for ray $r_i$ to get $n_{r,i}$ hits.
        \STATE Sample $N$ points with normalized location along $r_i$.
        \STATE Calculate volumetric rendering color $\hat{\boldsymbol{C_{r_i}}}$.
        \STATE Recording $w_i$ and $f(x_i)$ in to grid buffer at $x_i$.
    \ENDFOR 
\ENDFOR
\STATE Normalize the buffer for $\vec{w_t}$ and $\vec{f(x_t)}$.
\STATE Predict surface points $\vec{x_q}$ given queries $\vec{x_i}$.
\STATE Calculate $\mathcal{L}_{geo}(\vec{x_q}, \vec{x_t})$ and other losses.
\STATE Update network parameters via optimizer.
\STATE Refresh buffer with $\vec{w_{t}}$ and $f(\vec{x_{t}})$. 
\end{algorithmic}
\end{algorithm}
\section{Discussion on mult-view image annotation}
\begin{figure}[t!]
    \centering
    \includegraphics[width=\linewidth]{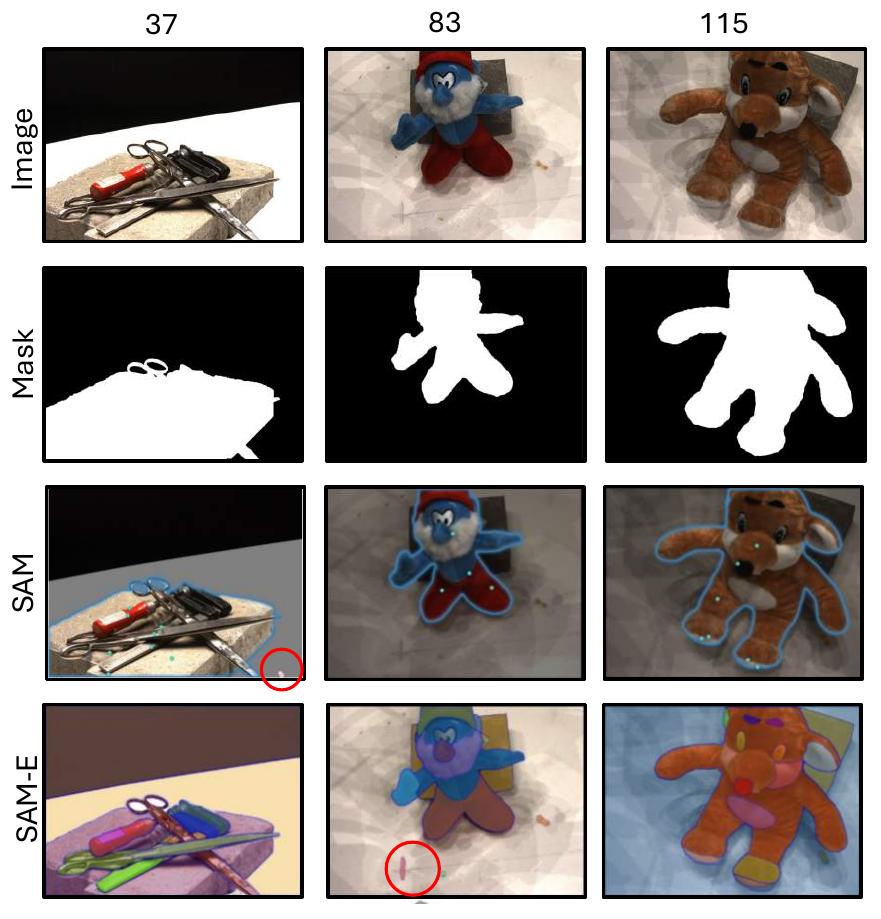}
    \caption{\textbf{Object masks for scenes in DTU.} We denote the scene number above.
    SAM and SAM-E are extracted mask results by hover \& click and everything modes, respectively. The artifacts of SAM predictions are highlighted by red circles.
    }
    \label{fig:masks}
\end{figure}

As shown in \cref{fig:masks}, the object masks vary significantly in their level of detail. For instance, Scene 37 requires intricate binary segmentation for elements like scissors with thin edges. Each scene in the DTU dataset includes either 49 or 64 images, making the per-scene annotation process extremely labor-intensive to achieve precise segmentation. Therefore, manual annotation becomes cumbersome in this context.
Additionally, Scenes 83 and 115 lack detailed masks for the bricks supporting toys, which affects the accuracy of performance evaluation on recovered bricks. This omission highlights the need for an approach like Hi-NeuS to handle segmentation without relying on foreground masks. Obtaining such masks is not only cumbersome but also poses challenges even for advanced models like the Segment Anything Model (SAM). For instance, the SAM and SAM-E results in \cref{fig:masks} show noticeable artifacts, which may pose issues for maintaining multi-view consistency when filtering meshes.
The complexity and effort required for manual annotation in datasets like DTU have been documented in various studies. For example, \cite{xie2018accurate} discusses the difficulties in achieving accurate segmentation in large-scale datasets due to high annotation costs and the time-consuming nature of the process. Similarly, \cite{kirillov2019panoptic} highlights the limitations of automatic segmentation models when dealing with fine-grained details in objects like thin edges and intricate shapes.
\begin{figure*}[t!]
    \centering
    \includegraphics[width=\linewidth]{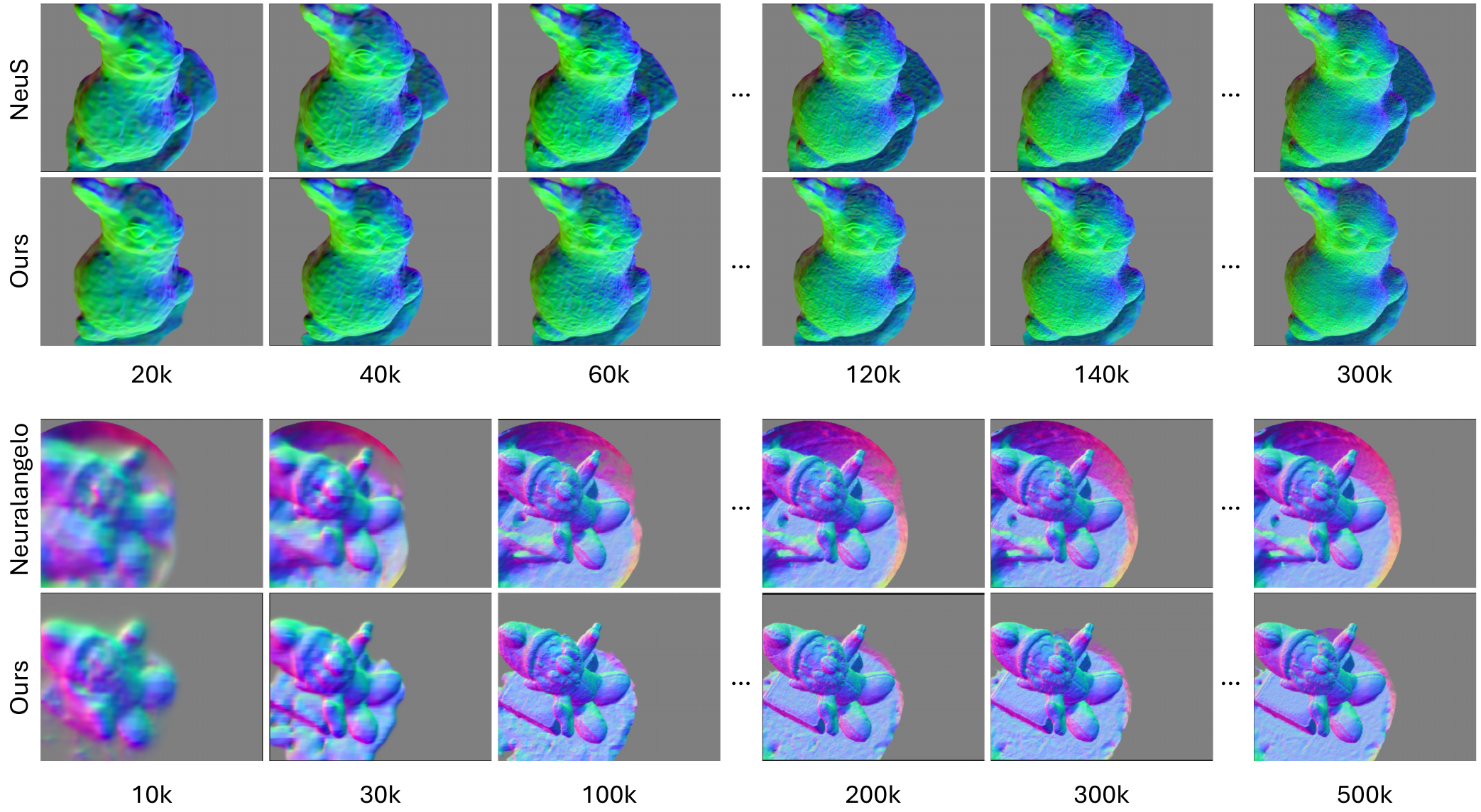}
    \caption{\textbf{The quality comparison between Hi-NeuS and NeuS based on different baselines during Training.}
    }
    \label{fig:converge}
\end{figure*}
By addressing these challenges, Hi-NeuS aims to provide a more efficient solution for object masking in complex scenes without the need for exhaustive manual annotation or heavy reliance on pre-existing foreground masks.

\section{Performance}

\begin{table*}[t!]
  \centering
  \caption{Quantitative results on DTU dataset~\cite{Jensen_Dahl_Vogiatzis_Tola_Aanaes_2014}. Proposed method consistently boosts the performance on the NeuS and Neuralangelo. }
  \label{tab:dtu_all}
  \resizebox{\linewidth}{!}{ 
    \begin{threeparttable}
    \begin{tabular}{llcccccccccccccccc} 
      \toprule[1.5pt]
      & & 24 & 37 & 40 & 55 & 63 & 65 & 69 & 83 & 97 & 105 & 106 & 110 & 114 & 118 & 122 & Mean \\ 
      \cline{2-18}
      \multirow{6}{*}{\rotatebox[origin=c]{90}{ unmasked CD ↓}} 
      & NeuS~\cite{Wang_Liu_Liu_Theobalt_Komura_Wang_2021} & 1.59 & 1.98 & 1.44 & 0.95 & 1.82 & 0.74 & 0.64 & 1.63 & 1.30 & 1.41 & 0.59 & 1.33 & 0.44 & 0.51 & 0.54 & 1.13 \\
      & Hi-NeuS(NeuS)-32  & \cellcolor{red!25}0.96 & \cellcolor{red!25}0.93 & 0.76 & 0.78 & 1.43 & \cellcolor{red!25}0.71 & 0.66 & 1.90 & \cellcolor{red!25}1.02 & 1.18 & \cellcolor{red!25}0.58 & 2.00 & 0.45 & 0.52 & 0.56 & 0.96 \\
      & Hi-NeuS(NeuS)-64 & 0.97 & 0.95 & \cellcolor{red!25}0.71 & 0.56 & 1.37 & 0.71 & 1.02 & \cellcolor{red!25}1.45 & 1.05 & 1.11 & 0.59 & \cellcolor{red!25}1.27 & 0.44 & 0.51 & 0.55 & 0.88 \\
      & Hi-NeuS(NeuS)-128 & 0.95 & 1.10 & 1.07 & \cellcolor{red!25}0.47 & \cellcolor{red!25}1.37 & 0.68 & \cellcolor{blue!25}0.66 & 1.47 & 1.25 & \cellcolor{red!25}1.07 & 0.60 & 3.86 & \cellcolor{red!25}0.44 & \cellcolor{red!25}0.50 & \cellcolor{red!25}0.54 & 1.07 \\
      \cline{2-18}
            & Gaussian Surfels~\cite{dai2024highquality} &1.00 & 1.97 & 1.06 & 1.74 & 2.32 & 2.35 & 2.02 & 3.48 & 2.45 & 2.55 & 2.31 & 8.13 & 1.49 & 2.69 & 3.48 & 2.60\\
      \cline{2-18}
      & Neuralangelo~\cite{Neuralangelo} &0.62 & 1.63 & 0.66 & 0.56 & 1.51 & 1.38 & 2.60 & 2.03 & 2.15 & 1.11 & 0.46 & 1.31 & 0.48 & 0.95 & 1.25 & 1.25 \\
      & Hi-NeuS(Neuralangelo)-64 & \cellcolor{red!25}0.55 & \cellcolor{red!25}1.55 & \cellcolor{red!25}0.61 & \cellcolor{blue!25}0.60 & \cellcolor{red!25}1.51 & \cellcolor{red!25}0.77 & \cellcolor{red!25}2.25 & \cellcolor{red!25}1.19 & \cellcolor{red!25}1.52 & \cellcolor{red!25}1.09 & \cellcolor{red!25}0.43 & \cellcolor{red!25}1.20 & \cellcolor{red!25}0.43 & \cellcolor{red!25}0.88 & \cellcolor{blue!25}1.36 & 1.06 \\
      \midrule[1.2pt]
      \multirow{10}{*}{\rotatebox[origin=c]{90}{ masked CD ↓}} 
      & NeRF~\cite{Mildenhall_Srinivasan_Tancik_Barron_Ramamoorthi_Ng_2020} & 1.90 & 1.60 & 1.85 & 0.58 & 2.28 & 1.27 & 1.47 & 1.67 & 2.05 & 1.07 & 0.88 & 2.53 & 1.06 & 1.15 & 0.96 & 1.49 \\
      & VolSDF~\cite{Yariv_Gu_Kasten_Lipman_2021} & 1.14 & 1.26 & 0.81 & 0.49 & 1.25 & 0.70 & 0.72 & 1.29 & 1.18 & 0.70 & 0.66 & 1.08 & 0.42 & 0.61 & 0.55 & 0.86 \\
      & RegSDF\textdagger~\cite{Zhang_Yao_Li_Fang_McKinnon_Tsin_Quan_2022} & 0.60 & 1.41 & 0.64 & 0.43 & 1.34 & 0.62 & 0.60 & 0.90 & 0.92 & 1.02 & 0.60 & 0.59 & 0.30 & 0.41 & 0.39 & 0.72 \\
      & NeuralWarp\textdagger~\cite{Darmon_Bascle_Devaux_Monasse_Aubry_2022} & 0.49 & 0.71 & 0.38 & 0.38 & 0.79 & 0.81 & 0.82 & 1.20 & 1.06 & 0.68 & 0.66 & 0.74 & 0.41 & 0.63 & 0.51 & 0.68 \\
      & D-NeuS\textdagger~\cite{Chen_Zhang_Feldmann_Schreer_Eisert_2022} &0.44 & 0.79 &0.35 & 0.39 &0.88 &0.58 &0.55 &1.35 & 0.91 &0.76 & 0.40 & 0.72 & 0.31  &0.39 & 0.39 & 0.61\\
      \cline{2-18}
      & NeuS~\cite{Wang_Liu_Liu_Theobalt_Komura_Wang_2021} &0.93 &1.07 &0.81 &0.38 &1.02 &0.60 &0.58 &1.42 &1.15 &0.78 &0.57 &1.16 &0.35 &0.45 &0.46 &0.78 \\
      & Hi-NeuS(NeuS)-32  & \cellcolor{red!25}0.77 & \cellcolor{red!25}0.90 & 0.73 & 0.37 & 1.00 & \cellcolor{red!25}0.59 & 0.59 & 1.42 & \cellcolor{blue!25}1.19 & 0.79 & \cellcolor{red!25}0.56 & 1.93 & 0.35 & 0.45 & 0.48 & 0.81\\
      & Hi-NeuS(NeuS)-64 &0.85 &0.92 &\cellcolor{red!25}0.68 &0.38 &1.09 &0.57 &0.65 &\cellcolor{red!25}1.40 &1.21 &0.80 &0.57 &\cellcolor{blue!25}1.11 &0.34 &0.44 &0.47 &0.77 \\
      & Hi-NeuS(NeuS)-128 & 0.76 & 1.07 & 0.95 & \cellcolor{red!25}0.37 & \cellcolor{red!25}0.99 & 0.56 & \cellcolor{blue!25}0.59 & 1.45 & 1.25 & \cellcolor{blue!25}0.89 & 0.59 & 3.78 & \cellcolor{red!25}0.33 & \cellcolor{red!25}0.45 & \cellcolor{red!25}0.46 & 0.98\\
      \cline{2-18}
      & Neuralangelo~\cite{Neuralangelo} & 0.39 & 0.72 & 0.35 & 0.33 & 0.82 & 0.74 & 1.70 & 1.34 & 1.95 & 0.71 & 0.47 & 1.00 & 0.33 & 0.82 & 0.78 & 0.83 \\      
      & Hi-NeuS(Neuralangelo)-64 &\cellcolor{red!25}0.39 &\cellcolor{red!25}0.71 &\cellcolor{blue!25}0.36 &\cellcolor{red!25}0.33 &\cellcolor{blue!25}0.92 &\cellcolor{red!25}0.55 &\cellcolor{red!25}1.42 &\cellcolor{red!25}1.25 &\cellcolor{red!25}1.44 &\cellcolor{blue!25}0.73 &\cellcolor{red!25}0.45 &\cellcolor{red!25}0.99 &\cellcolor{red!25}0.33 &\cellcolor{red!25}0.70 &\cellcolor{red!25}0.73 &0.75 \\
      \midrule[1.2pt]
      \multirow{9}{*}{\rotatebox[origin=c]{90}{PSNR ↑}} 
      & 
      RegSDF\textdagger~\cite{Zhang_Yao_Li_Fang_McKinnon_Tsin_Quan_2022} &24.78 &25.31 &23.47 &23.06 &22.21 &28.57 &25.53 &21.81 &28.89 &26.81 &27.91 &24.71 &25.13 &26.84 &21.67 &28.25  \\
      & VolSDF~\cite{Yariv_Gu_Kasten_Lipman_2021} & 26.28 & 25.61 & 26.55 & 26.76 & 31.57 &31.50 &29.38& 33.23 &28.03 &32.13& 33.16& 31.49& 30.33& 34.90& 34.75& 30.38  \\
      & NeRF~\cite{Mildenhall_Srinivasan_Tancik_Barron_Ramamoorthi_Ng_2020} & 26.24 & 25.74 & 26.79 & 27.57 & 31.96 & 31.50 & 29.58 & 32.78 &28.35 & 32.08 & 33.49 & 31.54 & 31.00 & 35.59 & 35.51 & 30.65 \\
      \cline{2-18}
      & 
NeuS~\cite{Wang_Liu_Liu_Theobalt_Komura_Wang_2021} &25.82 &23.64 &26.64 &25.60 &27.68 &30.83 &27.68 &34.04 &26.61 &31.35 &29.29 &28.08 &28.55 & 31.28 &33.68 &28.79 \\
      & Hi-NeuS(NeuS)-32 & \cellcolor{red!25}26.24 & \cellcolor{red!25}23.79 & 26.98 & 25.70 & 30.21 & \cellcolor{red!25}31.65 & 29.27 & 34.94 & \cellcolor{blue!25}26.59 & 32.31 & \cellcolor{red!25}32.37 & 29.30 & 28.73 & 34.15 & 33.69 & 29.73 \\
      & Hi-NeuS(NeuS)-64&26.25 &23.76 &\cellcolor{red!25}26.88 &25.63 &30.50 &31.57 &29.14 &\cellcolor{red!25}34.90 &26.55 &32.27 &32.27 &\cellcolor{red!25}29.43 &28.83 &34.00 &33.89 &29.72 \\
      & Hi-NeuS(NeuS)-128 & 26.14 & 23.56 & 26.90 & \cellcolor{blue!25}25.48 & \cellcolor{red!25}30.22 & 31.38 & \cellcolor{red!25}29.23 & 35.06 & 26.65 & \cellcolor{red!25}32.56 & 31.89 & 24.30 & \cellcolor{red!25}28.86 & \cellcolor{red!25}34.02 & \cellcolor{red!25}34.08 & 29.66 \\
      \cline{2-18}
      & Neuralangelo~\cite{Neuralangelo} & 30.90 & 28.01 & 31.60 & 34.18 & 36.15 & 36.30 & 34.10 & 38.84 & 31.28 & 37.15 & 35.73 & 33.60 &31.80 & 38.19 & 38.42 & 34.13 \\   
      & Hi-NeuS(Neuralangelo)-64 &\cellcolor{blue!25}30.80 &\cellcolor{red!25}28.01 &\cellcolor{blue!25}31.50 &\cellcolor{blue!25}29.82 &\cellcolor{blue!25}36.12 &\cellcolor{blue!25}36.17 &\cellcolor{blue!25}34.06 & \cellcolor{red!25}39.04 &\cellcolor{blue!25}31.13 &\cellcolor{red!25}37.18 &\cellcolor{blue!25}35.62 &\cellcolor{red!25}33.71 &\cellcolor{blue!25}31.53 &\cellcolor{blue!25}38.01 &\cellcolor{blue!25}38.07 &34.05 \\
      \midrule[1.2pt]
      \multirow{6}{*}{\rotatebox[origin=c]{90}{Noise\% ↓}} & 
NeuS~\cite{Mildenhall_Srinivasan_Tancik_Barron_Ramamoorthi_Ng_2020}&40.75 &60.50 &56.83 &72.60 &32.27 &28.69 &26.07 &75.41 &43.14 &64.46 &57.33 &17.35 &15.47 &8.53 &11.03 &39.13 \\
         & Hi-NeuS(NeuS)-32& \cellcolor{red!25}34.02 & \cellcolor{red!25}3.74 & 5.90 & 49.12 & 27.58 & \cellcolor{blue!25}29.52 & 22.04 & 67.33 & \cellcolor{red!25}26.98 & 61.79 & \cellcolor{red!25}32.90 & 19.47 & 14.71 & 17.10 & 15.33 & 28.50\\
      & Hi-NeuS(NeuS)-64 &34.02 &4.18 &\cellcolor{red!25}8.47 &33.06 &32.06 &30.76 &43.08 &\cellcolor{red!25}66.89 &29.13 &61.40 &32.63 &\cellcolor{blue!25}17.45 &14.44 &17.75 &17.19 &29.50 \\
            & Hi-NeuS(NeuS)-128 & 45.72 & 3.39 & 5.23 & \cellcolor{red!25}15.61 & \cellcolor{red!25}25.56 & 34.72 & \cellcolor{red!25}10.76 & 65.59 & 34.78 & \cellcolor{red!25}53.71 & 32.41 & 15.34 & \cellcolor{red!25}14.31 & \cellcolor{red!25}5.52 & \cellcolor{blue!25}12.94 & 24.90\\
      \cline{2-18}
                  & Gaussian Surfels~\cite{dai2024highquality} & 43.09 & 45.46 & 50.04 & 61.64 & 25.07 & 60.11 & 58.98 & 62.56 & 54.89 & 56.93 & 75.41 & 99.69 & 77.05 & 74.77 & 84.89 & 62.04 \\
      \cline{2-18}
  & Neuralangelo~\cite{Neuralangelo}&36.24 &52.32 &55.62 &66.63 &56.77 &57.84 &77.97 &76.70 &57.71 &63.60 &39.52 &84.71 &49.13 &35.34 &51.41 & 57.44  \\
      & Hi-NeuS(Neuralangelo)-64 &\cellcolor{red!25}32.36 &\cellcolor{red!25}44.25 &\cellcolor{red!25}39.16 &\cellcolor{red!25}59.96 &\cellcolor{red!25}43.01 &\cellcolor{red!25}34.23 &\cellcolor{red!25}68.31 &\cellcolor{red!25}61.89 &\cellcolor{blue!25}58.68 &\cellcolor{red!25}60.71 &\cellcolor{red!25}36.15 &\cellcolor{red!25}17.45 &\cellcolor{red!25}21.54 &\cellcolor{red!25}28.42 &\cellcolor{blue!25}57.60 &45.67 \\
      \bottomrule[1.5pt]
    \end{tabular}%
    \begin{tablenotes}
\item * 
%
\textdagger \ denotes auxiliary data inputs, including 3D points from SFM or other pretrained models.
We denote our models as Hi-NeuS(backbone)-grid resolution, with selected variants highlighted. Compared to the baselines, our models demonstrate superior performance, highlighted in \textcolor{red}{red}, while the sub-optimal is marked in \textcolor{blue}{blue} for each measure and scene.
\end{tablenotes}
\end{threeparttable}
}
\end{table*}

\begin{figure*}[t!]
    \centering
    \includegraphics[width=0.85\linewidth]{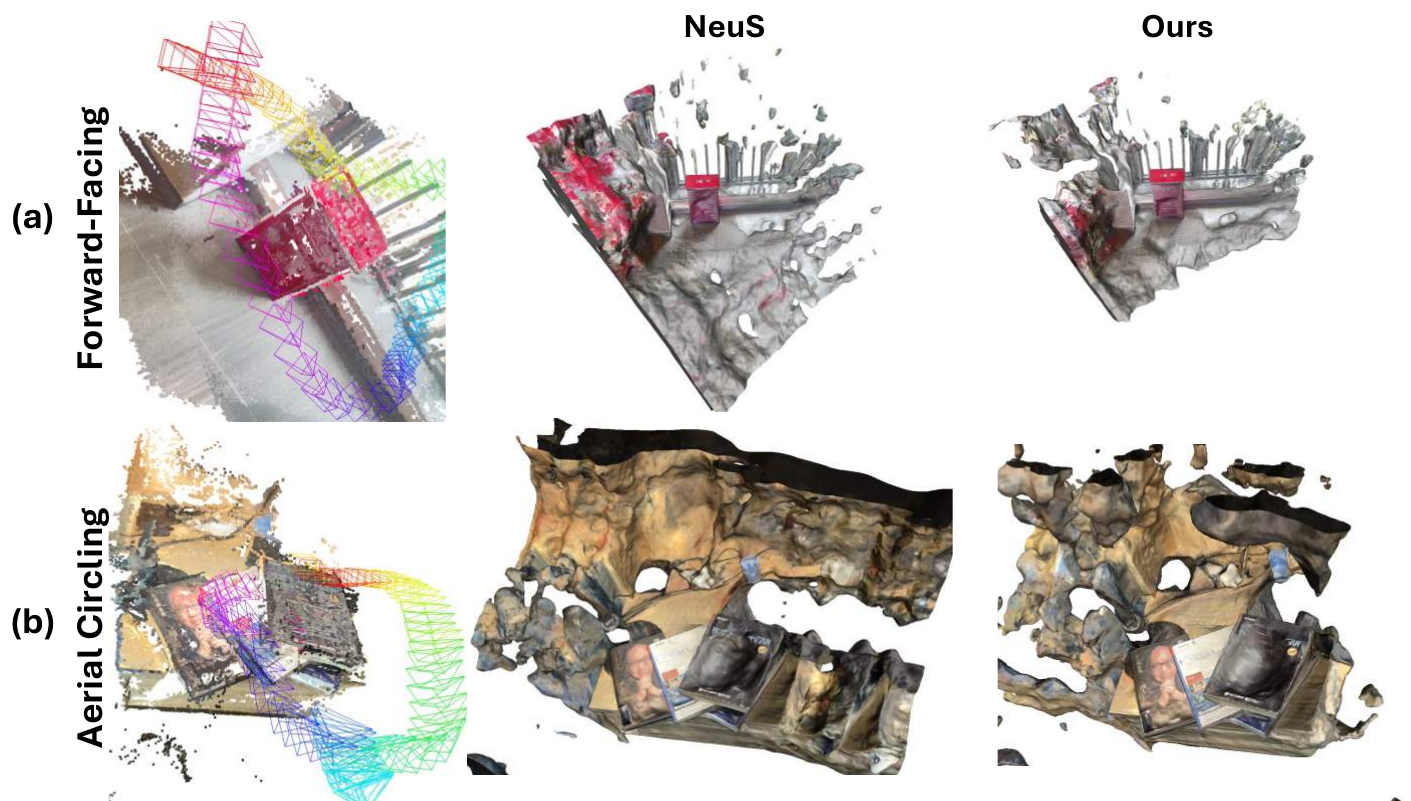}
    \caption{\textbf{More challenging real-world capturing.} The first column depicts the COLMAP results with our predicted camera poses. The other two columns compare the reconstructed scenes with the same resolution and camera viewpoints on their quality. 
    }
    \label{fig:cap_types}
\end{figure*}
%

\subsection{Additional ablation studies}
\noindent\textbf{Training quality and convergence. }
In \cref{fig:converge}, we show the norm maps during the training process of Hi-NeuS and compare them with NeuS. We observe that the compact result appears at a very early stage of training, for example, at 20k/300k or 10k/500k iterations. This indicates that Hi-NeuS’s SDF representation remains compact, focusing on objects rather than surrounding noise, which demonstrates improved geometry accuracy at earlier stages. Throughout the training process, we maintain this compactness, whereas our baselines tend to accumulate noise, potentially due to uncertainty accumulation. Notice that Neuralangelo has more noise in both the surroundings and objects. In contrast, Hi-NeuS successfully produces a more compact structure, achieving compactness and geometry accuracy throughout training.

\subsection{Overall performance on the selected model variants}
\begin{figure*}[t!]
    \centering
    \includegraphics[width=\linewidth]{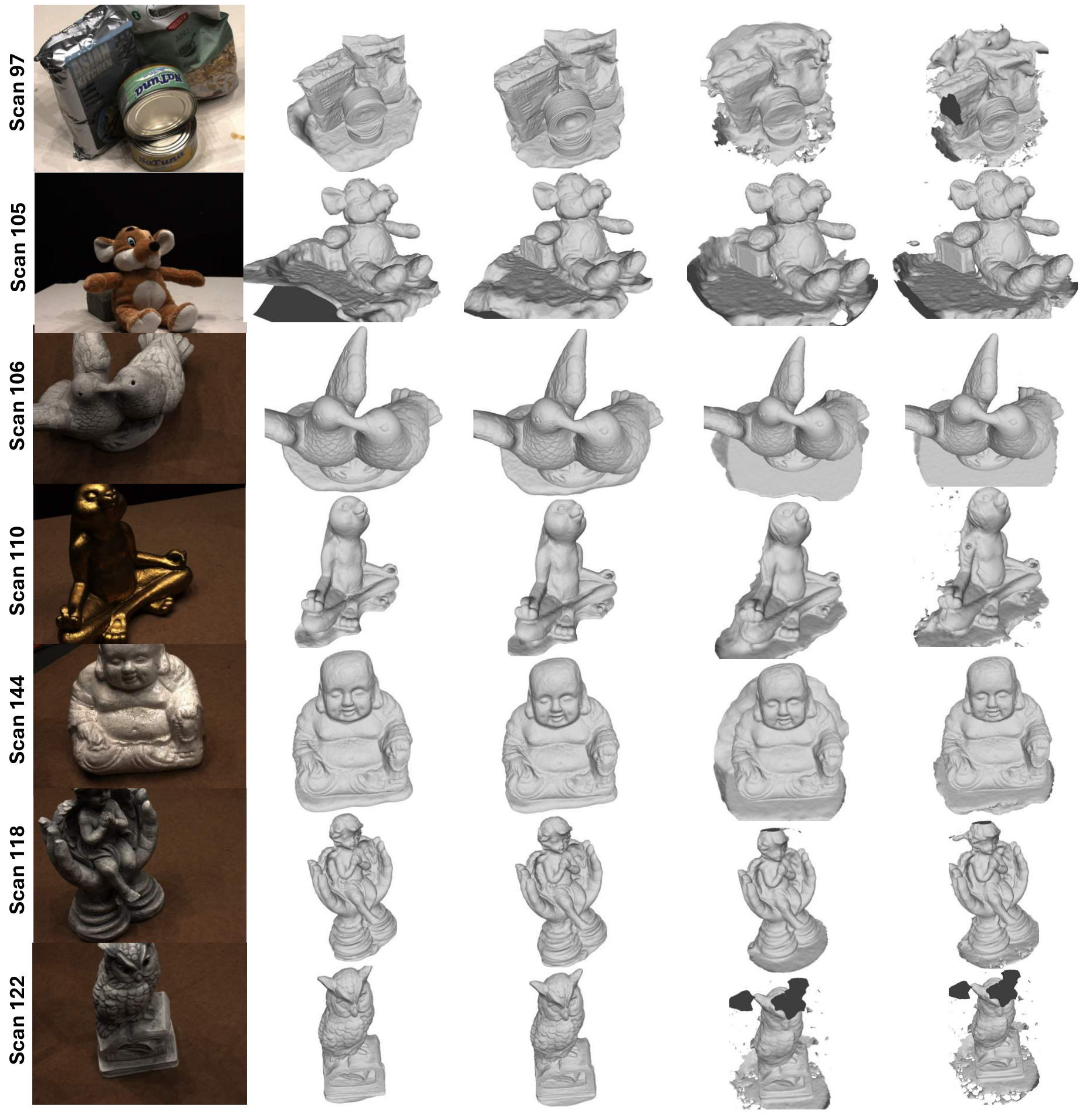}
    \caption{\textbf{The full evaluation result on the DTU dataset.}
    }
    \label{fig:dtu_full}
\end{figure*}

\begin{figure*}[t!]
    \centering
    \includegraphics[width=0.75\linewidth]{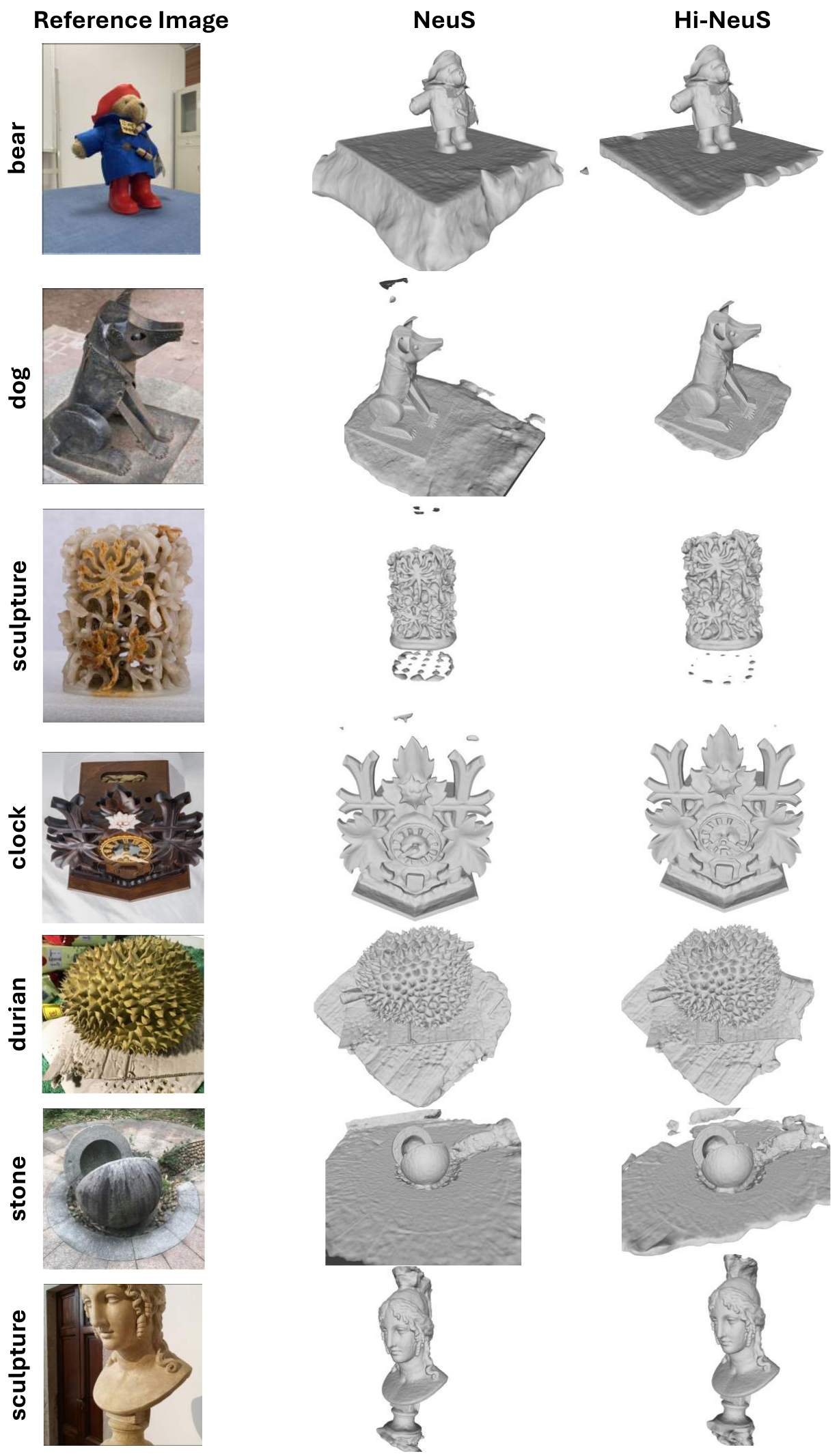}
    \caption{\textbf{The full evaluation result on the BlendedMVS dataset.}
    }
    \label{fig:blend_full}
\end{figure*}

In \cref{tab:dtu_all}, we identify the optimal model variants across different grid resolutions, where all selected variants are highlighted for each scene. 
During our model selection, we prioritize the ones with less mesh noise rather than geometric accuracy and rendering quality. 
In \cref{fig:dtu_full} and \cref{fig:blend_full}, we list all uncolored mesh results for NeuS and Neuralangelo. 

\subsection{Training cost}
We use an NVIDIA A800-SXM4-40GB GPU to evaluate the training cost, averaging the results on the DTU dataset. For NeuS, we report memory consumption as follows: 10.25GB for a resolution of $128\times128\times128$, 9.58GB for $64\times64\times64$, 9.03GB for $32\times32\times32$, and 8.23GB for our NeuS baseline. For Neuralangelo, the memory cost at a resolution of $64\times64\times64$ is 19.47GB, compared to its baseline of 18.98GB. The inference speed is 0.08 seconds per iteration, compared to the original 0.05 seconds per iteration.

\subsection{More Challenging Real-world Capturing}
To assess our model's capability in surface reconstruction in real-world scenarios,
where videos may not cover sufficient views as people prefer to shoot videos while walking freely.
We categorize human capturing scenarios into two main types, excluding the object-centric approach:

\textbf{(1) Forward-Facing}: Camera poses primarily focus on the front parts of objects, leaving back regions under-explored.

\textbf{(2) Aerial Circling}: When cameras are placed above objects, views are mostly concentrated on the upper regions, potentially neglecting the bottom and side views.

As illustrated in \cref{fig:cap_types}, we evaluate the performance of Hi-NeuS in comparison to its NeuS backbone.
The results show that the reconstructed scenes exhibit improved compactness with reduced artifacts.
This suggests that similar to the object-centric approach, our method can focus on the regions that capture most of the overlaps from diverse view perspectives.
This can align with users' intentions on the region of interest and mitigate geometry bias, particularly when predicted camera poses are not sufficiently accurate.

\begin{figure*}[t!]
    \centering
    \includegraphics[width=\linewidth]{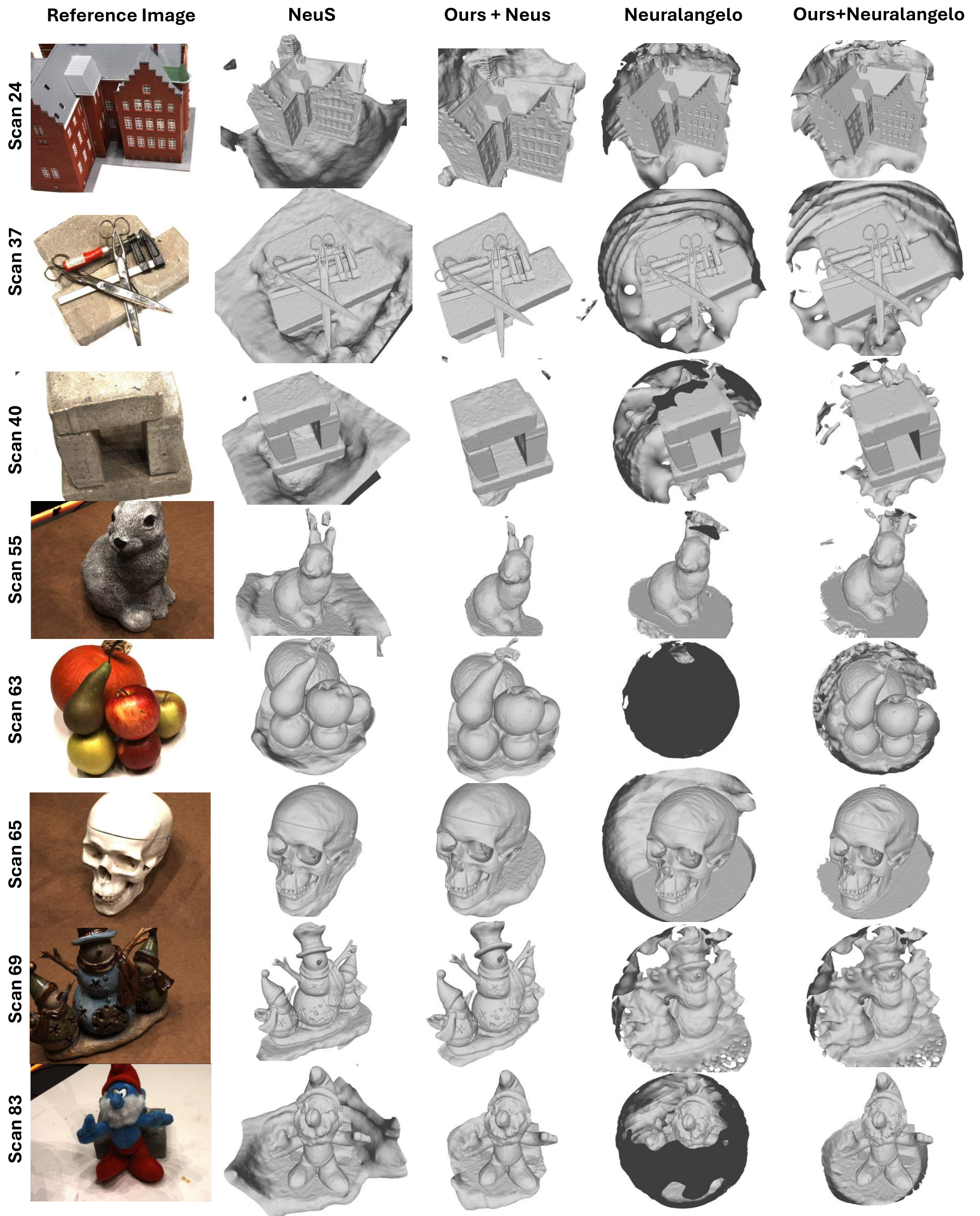}
\end{figure*}

\section{Video demo}
We attach the video demo of our full capturing and reconstruction process, the key idea illustration, and visualization with the quality comparison with the rotating camera views.
\clearpage


\end{document}